\documentclass[11pt]{article}

\usepackage[final]{acl}

\usepackage{times}
\usepackage{latexsym}
\usepackage{amsmath}
\usepackage{subcaption}
\usepackage{booktabs}
\usepackage{multirow}

\usepackage{graphicx}
\usepackage{amsmath}
\usepackage{amssymb}
\usepackage{mathtools}
\usepackage{subcaption}
\usepackage{algorithm}
\usepackage[most]{tcolorbox}

\newtcolorbox{outputbox}[2][]{enhanced, breakable,
  colback=cyan!8,            
  colframe=cyan!55!black,    
  fonttitle=\bfseries,
  title={#2},
  coltitle=black,
  top=5pt, bottom=5pt, left=7pt, right=7pt,
  boxrule=0.4pt,
  sharp corners,
  #1
}

\newtcolorbox{memorybox}[1][]{enhanced, breakable, colback=orange!5!white, colframe=orange!50!black, title=#1, boxrule=0.6pt, arc=2pt, left=6pt, right=6pt, top=6pt, bottom=6pt}

\newtcolorbox{decisionbox}[1][]{enhanced, breakable,
colback=blue!5!white, colframe=blue!65!black,
fonttitle=\bfseries, coltitle=black, title=#1,
boxrule=0.7pt, arc=3pt, left=8pt, right=8pt, top=6pt, bottom=6pt,
attach title to upper, before skip=6pt, after skip=8pt}

\newtcolorbox{memoryopbox}[1][]{enhanced, breakable,
colback=green!5!white, colframe=green!60!black,
fonttitle=\bfseries, coltitle=black, title=#1,
boxrule=0.7pt, arc=3pt, left=8pt, right=8pt, top=6pt, bottom=6pt,
attach title to upper, before skip=6pt, after skip=8pt}

\tcbset{
  promptbox/.style={
    enhanced,
    breakable,            
    colback=white,        
    colframe=black,       
    boxrule=0.5pt,        
    arc=2pt,              
    outer arc=2pt,
    fonttitle=\bfseries,  
    coltitle=white,       
    colbacktitle=black,   
    title=Prompt,         
    attach boxed title to top left={
      yshift=-1mm, xshift=1mm
    },
    boxed title style={
      sharp corners,
      boxrule=0pt,
      colback=black,
    },
    top=2mm, bottom=2mm, left=2mm, right=2mm,
    before skip=4pt, after skip=4pt,    
  }
}
\usepackage{cuted} 

\usepackage{bm}
\usepackage{colortbl}
\arrayrulecolor{black}

\definecolor{ourworkbg}{RGB}{230, 246, 255}
\definecolor{blockbg}{RGB}{243,246,249}
\definecolor{bestcolor}{HTML}{E64B75}
\definecolor{secondbestcolor}{HTML}{009999}

\usepackage[T1]{fontenc}

\usepackage[utf8]{inputenc}

\usepackage{microtype}

\usepackage{inconsolata}
\usepackage{hyperref}
\usepackage{graphicx}

\title{Agent Memory with Episodic Retrieval for Financial Decision-Making}

\author{
 \textbf{Nuoyue Xu\textsuperscript{1,*}},
 \textbf{Jiang Liu\textsuperscript{2,*}},
 \textbf{Wenxuan Huang\textsuperscript{3,*}},
 \textbf{Xiang Zhang\textsuperscript{1}},
 \textbf{Juntai Cao\textsuperscript{1}},
  \textbf{Jiaqi Wei\textsuperscript{4}}
\\
\\
 \textsuperscript{1}University of British Columbia,
 \textsuperscript{2}University of Science and Technology of China,\\
 \textsuperscript{3}Fudan University,
 \textsuperscript{4}Zhejiang University
\\
 \textsuperscript{*}Equal contribution.}

\begin{document}
\maketitle

\begin{figure*}[t]
    \centering
    \includegraphics[width=\textwidth]{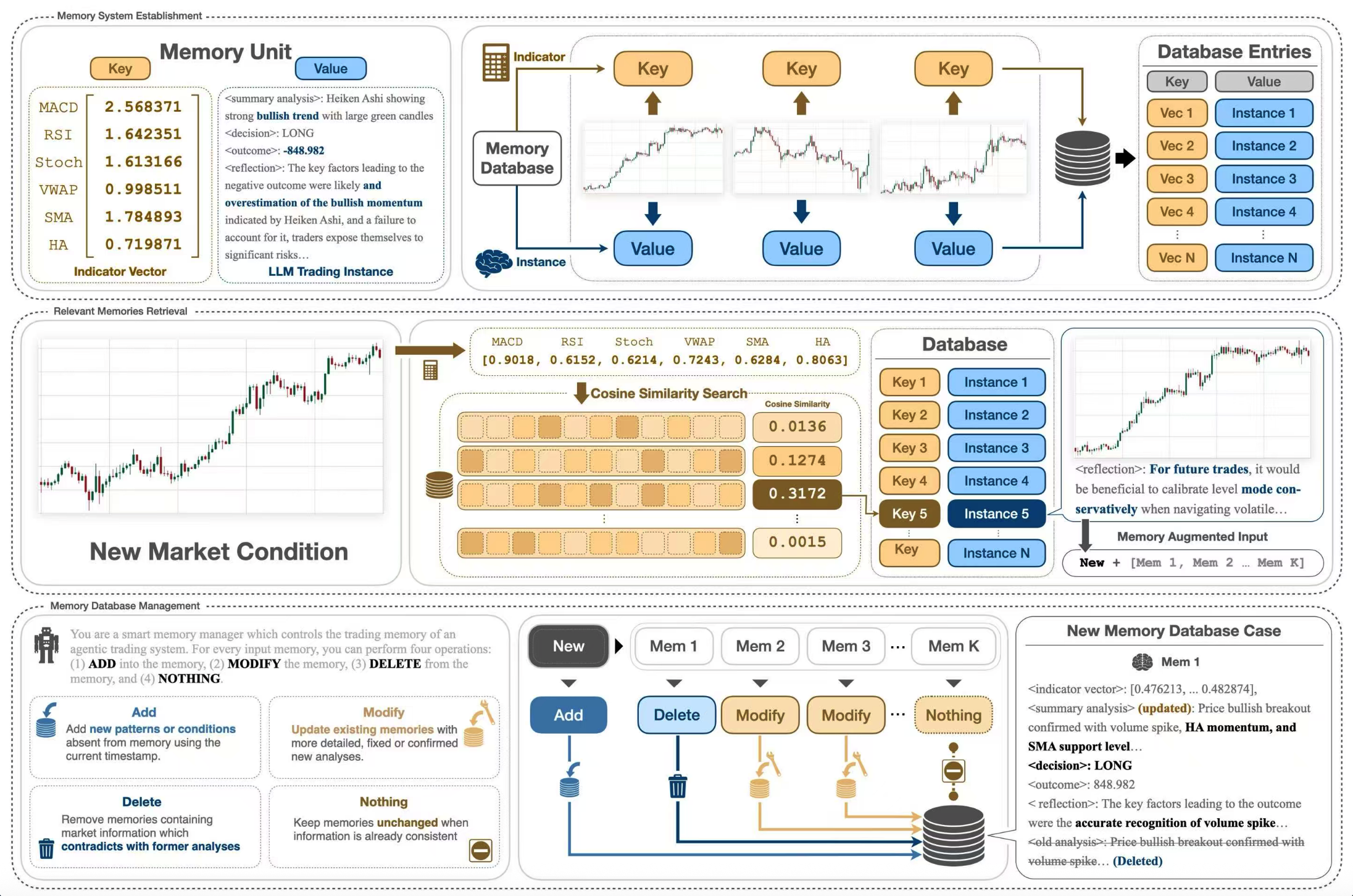}
    \caption{\textbf{Architecture of the \textsc{META} memory system.}
    The framework consists of three core components: Memory System
    Establishment, Relevant Memory Retrieval, and Memory Database Management.}
    \label{fig:meta_memory_system}
\end{figure*}

\begin{abstract}
Large language models (LLMs) have demonstrated strong capabilities in financial analysis and reasoning, inspiring recent advances in agent-based trading frameworks. While these systems show promise, prior approaches either emphasize long-horizon forecasting or operate as stateless analyzers, limiting their applicability to the demands of trading in complicated settings. To address these gaps, we introduce META (Memory Enhanced Trading Agent), the first RAG-like episodic-memory-augmented multi-agent framework for financial decision making. META integrates a family of specialized indicator agents (e.g., Trend, MACD, Stochastic, RSI, SMA, AVWAP, Heikin-Ashi) with a Decision Agent that fuses their reports, and a Memory module that retrieves and updates past trading episodes encoded as market state embeddings with outcomes and reflections. By recalling relevant experiences and adaptively reweighting signals under similar market regimes, META achieves improved directional accuracy and robustness under short-horizon evaluation. Our results demonstrate that \textbf{episodic memory} provides a powerful mechanism for regime-aware, interpretable, and low-latency decision-making in trading and decision making. Code of this project is released at \href{https://github.com/KriXord/Agent-Memory-with-Episodic-Retrieval-for-Financial-Decision-Making}{GitHub}.

\end{abstract}

\section{Introduction}

Building autonomous agents capable of reliable decision-making in complex, non-stationary environments remains a central challenge in machine learning. High-frequency trading (HFT) exemplifies this difficulty: agents must process rapidly evolving, high-dimensional data under strict latency constraints, where profitable patterns can invert within minutes \citep{ yang2023fingpt, zhang2023instruct}. Traditional quantitative systems rely on handcrafted logic for low-latency execution but collapse under regime shifts. Conversely, Large Language Model (LLM) agents demonstrate advanced reasoning ability \citep{wang2024quantagent, li2023tradinggpt}, yet remain \emph{stateless}—each decision made in isolation, without learning from prior outcomes. This absence of experiential continuity limits adaptability in markets defined by recurrence and volatility.

We argue that progress in this domain requires agents endowed with \textbf{experiential memory}, capable of recognizing and reusing structural analogies between past and present market states \citep{yu2023finmem, ji2023towards}. While dynamic memory frameworks such as Mem0 and MemGPT \citep{chhikara2025mem0, packer2023memgpt, xu2025amem} enable long-term reasoning in text-based tasks, they fail to generalize to the stochastic, high-dimensional structure of financial time series. Unlike language, market data evolve as continuous numerical manifolds, demanding memory systems that operate over both symbolic abstractions and quantitative dynamics. This motivates a new direction: re-engineering episodic memory for real-time environments where state representations are not linguistic but numerical and non-stationary.

To this end, we introduce \textbf{META} (Memory Enhanced Trading Agent), a multi-agent cognitive architecture that reframes trading as a “Perceive–Recall–Synthesize” process. In the \textbf{Perception} stage, specialized Signal Agents extract interpretable technical features from market streams. In the \textbf{Recall} stage, a Memory Operator performs case-based retrieval from prior trades to identify historically analogous contexts. Finally, the \textbf{Synthesis} stage fuses real-time analysis with retrieved precedents to produce context-aware trading decisions. This process is powered by a \textbf{dual-stream state representation} that separates symbolic indicator vectors (for reasoning and retrieval) from normalized price–volume embeddings (for similarity search), balancing interpretability and numerical precision.

\begin{figure*}[t] \centering \includegraphics[width=\textwidth]{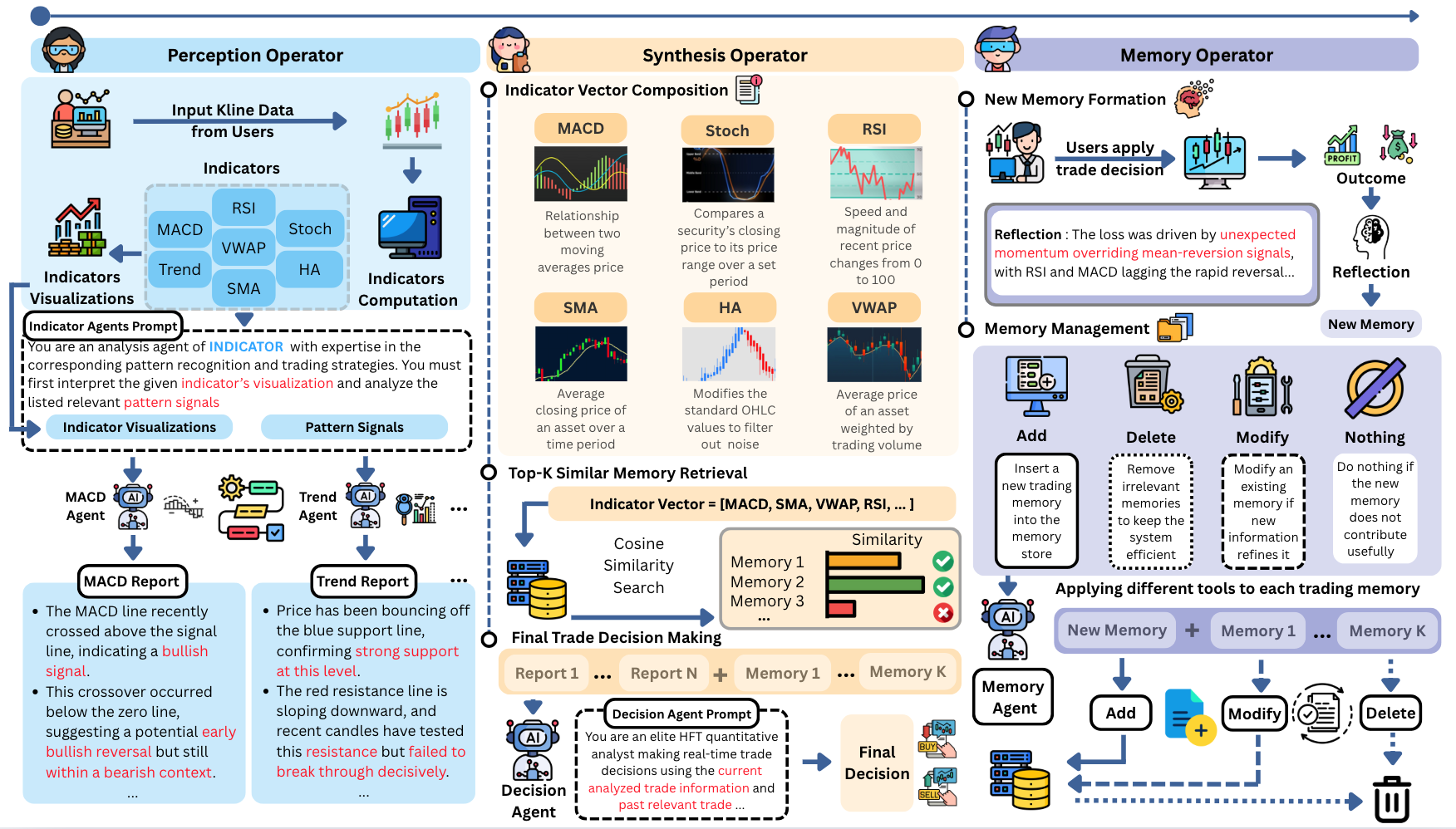} \caption{The META cognitive workflow, illustrating the interplay of its three core operators. The process unfolds in stages: (1) \textbf{Perception ($\Phi$)}: A distributed ensemble of Signal Agents encodes the market state in parallel. (2) \textbf{Synthesis ($\Pi$)}: Real-time reports are fused with retrieved historical precedents to formulate a decision. (3) \textbf{Memory ($\Psi$)}: The memory buffer is updated via a reflection-augmented consolidation cycle.} \label{fig:main} \end{figure*}

META is the first cognitive trading framework to embed retrieval-augmented episodic memory directly into the high-frequency decision loop. Its memory operator extends dynamic memory principles \citep{chhikara2025mem0, li2023compressing, an2025effective} to structured numerical domains, enabling adaptive reasoning grounded in market feedback. Empirically, META achieves substantial gains in profitability and stability over stateless baselines, showing that coupling cognition with memory yields agents that not only react to markets—but \emph{learn} from them.

\section{Related Work}
\noindent \paragraph{LLMs for Financial Decision-Making.}
Large Language Models (LLMs) are increasingly deployed as autonomous agents for financial trading, analysis, and alpha discovery \cite{yang2023fingpt, zhang2023instruct, wu2023bloomberggpt,gong2025multiprocessor,xiong2026quantharnesspricedrivenmultiagentllms}. LLM-based trading agents generally follow three paradigms: leveraging sentiment from news and social media \cite{lopez2023can, kirtac2024sentiment, zhang2024unveiling}, using reinforcement learning (RL) to refine policies through simulated trading \cite{ding2023integrating, koa2024learning}, or employing explicit reasoning frameworks \cite{ji2023towards, li2023tradinggpt, yu2023finmem, zhang2024multimodal,zhang2023don}. 
A complementary line of work formalizes such inference-time reasoning as an explicit search process over intermediate decision states \cite{wei2026llm}, and documents the broader transition from single-model prompting to autonomous, agentic systems that plan and act over extended horizons \cite{wei2025ai}.
Our work, META, advances the reasoning-driven paradigm with a specialized memory framework for technical analysis. It uses an ``Indicator Vector'' (e.g., MACD, RSI) to retrieve analogous historical scenarios via cosine similarity. A key contribution is our \textit{Memory Update and Reflection} loop, which refines memory based on trade outcomes, allowing the agent to learn from the performance of technical patterns. Unlike systems that generate new alpha factors \cite{xiong2025quantagent, wang2023alphagpt,zhang2025tokenization}, META focuses on adaptive decision-making with established indicators.

\noindent \paragraph{Memory Mechanisms in Agents.}
Long-term reasoning is a central challenge for LLM agents. Early methods that appended full interaction histories to prompts were inefficient and scaled poorly \cite{hatalis2023memory, liu2024lost, an2025effective}. This spurred the development of external memory frameworks, including layered architectures, dynamic knowledge networks, and cognitively inspired systems \cite{packer2023memgpt, xu2025amem, lee2024human, salama2025meminsight, liu2022character,wang2023selfcontrolled,zhao2025timeseriesscientist}.
However, persistent challenges include memory modules that are decoupled from the agent's reasoning policy \cite{li2023compressing, yoon2024compact, chhikara2025mem0} and RL-based systems that still rely on full-history accumulation without selective retention \cite{jin2025searchr1, zheng2025deepresearcher}. 
More recent retrieval-augmented systems further show that retrieval alone is insufficient: retrieved evidence must be critiqued and optimized at test time before it can reliably inform a decision \cite{wei2026retrieval}. META addresses these limitations in quantitative trading by tightly integrating a structured, indicator-based memory with a reflection-driven decision cycle. By using cosine similarity retrieval and continuously updating its knowledge based on trade performance, META enhances adaptivity in non-stationary market environments.




\section{Method}

We propose \textbf{META} (\textbf{M}emory-\textbf{E}nhanced \textbf{T}rading \textbf{A}gent), a cognitive framework that decomposes the trading process into three modular operators: Perception ($\Phi$), Memory ($\Psi$), and Synthesis ($\Pi$). Unlike conventional end-to-end black-box models, META introduces an explicit and auditable reasoning workflow. Its compositional design enables functional modularity, leverages non-parametric memory for generalization, and supports interpretable, retrieval-grounded decision making.

\subsection{Problem Formulation}

We cast trading as the problem of learning a \emph{retrieval-augmented policy} $\pi$, which maps a market state $z_t \in \mathcal{S}$ and a time-varying episodic memory $\mathcal{M}_t$ to a structured decision $d_t$. The objective is to learn an optimal policy $\pi^\star$ that maximizes the expected cumulative return:
\[
J(\pi) = \mathbb{E}\left[\sum_{t=0}^{T} \gamma^t\, o_t(d_t)\right],
\pi^\star = \arg\max_\pi J(\pi),
\]
where $o_t$ denotes the outcome from decision $d_t$, and $\gamma \in (0, 1]$ is a temporal discount factor.

At the heart of META lies the hypothesis that optimal trading policies are \emph{functionally decomposable}. We define $\pi$ as the composition of two interpretable modules:
\[
d_t = \Pi\big(\underbrace{\Phi(z_t)}_{\text{Perception}},\ \underbrace{\Psi(z_t, \mathcal{M}_t)}_{\text{Memory}}\big).
\]
This structured decomposition confers three main benefits: (i) \textbf{specialized processing} of heterogeneous financial inputs; (ii) \textbf{retrieval-augmented generalization} through episodic memory; and (iii) \textbf{transparent inference chains} that elucidate the agent’s decision logic (Fig.~\ref{fig:main}).

\subsection{Bifurcated State Representation}

Financial markets exhibit both high-level structural regimes and fine-grained temporal fluctuations. To model this duality, META employs a \emph{bifurcated state representation} that jointly captures symbolic abstractions and numerical microstructure. From a sliding window of observations $X_t = \{(O, H, L, C, V)_{t-i}\}_{i=0}^{W-1}$, we construct:
\[
z_t = \left[z^{(\mathrm{sym})}_t \, \big\| \, z^{(\mathrm{num})}_t\right],
\]
where $\|$ denotes concatenation.

The \textbf{symbolic component} $z^{(\mathrm{sym})}_t \in \mathbb{R}^M$ encodes several interpretable, domain-specific features—such as momentum, volatility, and mean-reversion indicators—using compact heuristic transformations. This channel supports logical and rule-based reasoning, providing the cognitive substrate for high-level pattern recognition and symbolic inference.

The \textbf{numerical component} $z^{(\mathrm{num})}_t \in \mathbb{R}^{5W}$ preserves the normalized, high-fidelity raw market microstructure (open, high, low, close, volume). It enables precise retrieval through vector similarity, capturing minute market configurations often lost in purely symbolic abstractions.

This bifurcated encoding allows the architecture to reason at two complementary levels: the symbolic stream offers interpretability and abstraction, while the numerical stream ensures precision and adaptability. By leveraging both, META bridges symbolic and subsymbolic reasoning.

\subsection{Perception Operator ($\Phi$)}

The perception operator $\Phi$ implements a distributed ensemble of \emph{Signal Agents} $\{A^m\}_{m=1}^M$, where each agent embodies a domain heuristic or learned detector (see Appendix~\ref{sec:indicator_computation} for the detailed mathematical formulation of each indicator). Each $A^m$ processes the market state $z_t$ to produce a compact report:
\[
r^m_t = A^m(z_t).
\]
The collective output forms a perceptual set $R_t = \{r^m_t\}_{m=1}^M$, in which each report $r^m_t$ encapsulates both quantitative metrics (e.g., indicator strength) and qualitative semantics (e.g., \textit{“bullish divergence,” “breakout detected”}). 

This design provides two crucial benefits. First, it modularizes perception—each agent can be refined or replaced independently, supporting continuous integration of new trading signals. Second, it yields structured, interpretable percepts, allowing higher modules to perform symbolic reasoning without operating directly on raw data. The same principle of decomposing a complex task across role-specialized LLM agents, rather than a single monolithic model, has proven effective in multi-agent generation pipelines across domains \cite{zhang2025postergen}, and recent work further learns to optimize the workflow over such agents at execution time \cite{xu2026evomas}.

\subsection{Memory Operator ($\Psi$)}

The memory operator $\Psi$ implements \emph{case-based analogical reasoning}. The episodic memory $\mathcal{M} = \{m_i\}_{i=1}^N$ stores discrete trading experiences, where each entry $m_i = (z_i, a_i, o_i, \rho_i)$ logs the state, action, outcome, and an optional textual reflection summarizing the context and lesson learned.

\paragraph{Retrieval.} 
Given the current state $z_t$, $\Psi$ identifies historically similar contexts by computing cosine similarity over the numerical embeddings:
\[
\mathrm{sim}(z_t, z_i) = \frac{(z_t^{(\mathrm{num})})^\top z_i^{(\mathrm{num})}}{\|z_t^{(\mathrm{num})}\| \|z_i^{(\mathrm{num})}\|}.
\]
A top-$k$ selection of analogous cases forms the retrieved neighborhood:
\[
\mathcal{N}_t = \{(m_i, s_i) \mid s_i = \mathrm{sim}(z_t, z_i) \ge \tau\}_{\text{Top-}k}.
\]
These retrieved cases constitute the agent’s “episodic evidence,” grounding its current reasoning in empirical precedent.

\paragraph{Consolidation.} 
To preserve computational efficiency and diversity, $\Psi$ employs a dynamic curation operator $\mathrm{Op}(m_t, \mathcal{M})$ that manages the lifecycle of memories. Low-utility or redundant entries are pruned, while novel, high-impact experiences are retained. This continual abstraction transforms $\mathcal{M}$ into a structured, evolving repository of archetypal scenarios—a non-parametric analog of synaptic consolidation in biological systems. 
Crucially, retrieved cases are not consumed verbatim: every episode carries the outcome and reflection of the original trade, so retrieved evidence is re-critiqued and re-weighted at test time instead of being naively appended to the prompt \cite{wei2026retrieval}.

\subsection{Synthesis Operator ($\Pi$)}

The synthesis operator $\Pi$ employs a LLM to integrate perceptual summaries $R_t$ and retrieved episodes $\mathcal{N}_t$ into a structured decision output:
\[
d_t \triangleq (a_t, \kappa_t, \mathcal{E}_t, \rho_t),
\]
where $a_t$ denotes the action (e.g., \texttt{LONG}), $\kappa_t$ is the confidence score, $\mathcal{E}_t \subseteq R_t \cup \mathcal{N}_t$ is the subset of selected evidence, and $\rho_t$ is a concise natural language rationale justifying $a_t$.

We enforce an \textbf{evidence-first, constrained decoding} process: the LLM must first select $\mathcal{E}_t$ from the available context, and only then generate the rationale $\rho_t$. This separation enforces grounded reasoning and produces an auditable explanation chain from input evidence to trading action. 
How such a reasoning chain is prompted is not a neutral design choice: the structure of the chain-of-thought prompt interacts with the answer space and determines whether deliberation helps or hurts \cite{zhang-etal-2025-prompt-design}. 
Likewise, deciding how much reasoning to expose as an explicit rationale is itself a policy to be learned, since verbose deliberation incurs latency while silent decisions are unauditable \cite{wei2026think}. 
META adopts a fixed, domain-motivated disclosure policy---select evidence first, then emit a short justification---which keeps the additional token budget small enough to remain within HFT latency limits.

\section{Experiments and Results}


\subsection{Experimental Setup}

To evaluate the multi-agent framework of META, we perform experiments on a asset consisting of five representative financial assets constructed by prior work \citet{xiong2025quantagent}: Crude Oil Futures (CL), S\&P 500 E-mini Futures (ES), Nasdaq-100 Futures (NQ), Invesco QQQ ETF (QQQ), and Bitcoin (BTC). Each dataset consists of 4-hour OHLCV (Open, High, Low, Close, Volume) sequences sampled over multi-year horizons, capturing diverse market regimes and volatility patterns. Following the experimental protocol of \citet{xiong2025quantagent}, each asset is partitioned into non-overlapping windows of 100 candlesticks, with the final three windows reserved for validation.

For the memory configuration, META employs the top-$k$ and similarity-threshold retrieval strategy. Unless otherwise stated, $k$ is set to 5 and the similarity threshold $\tau$ to 0.7. This configuration balances efficiency and contextual relevance for low-latency operation.

As baselines, we include the \textsc{QuantAgent} framework~\citep{xiong2025quantagent}—a multi-agent trading system without long-term memory—and a random decision generator that selects between \textsc{Long} and \textsc{Short} uniformly at random while sharing the same evaluation segments and reporting format.

\noindent\paragraph{Evaluation.}
We adopt the evaluation protocol proposed by \citet{xiong2025quantagent} to ensure comparability. The primary metric is \textbf{directional accuracy} ($\alpha$), which quantifies the correctness of \textsc{Long}/\textsc{Short} forecasts relative to subsequent price movements. In addition, we report four complementary return-based metrics from \citet{xiong2025quantagent} that jointly assess profitability, robustness, and memory relevance:
    Average cumulative return correlation ($R_{cc}$)
    Similarity-weighted return alignment ($R_{\text{sim}}$)
    Maximum realized gain ($R_{\max}$)
    Maximum drawdown ($R_{\min}$)
Detailed definitions, computation procedures, and theoretical motivation for these metrics are provided in Appendix~\ref{sec:metrics_design}.

\begin{table*}[t]
\centering
\renewcommand{\arraystretch}{0.95} 
\setlength{\tabcolsep}{8pt}        
\small
\resizebox{\textwidth}{!}{
\begin{tabular}{l|c| c c c c c}
\toprule
\textbf{Method} &
\textbf{Accuracy \bm{$\alpha$} (\% $\uparrow$)} &
\textbf{\bm{$\Delta\alpha$} (\% $\uparrow$)} &
\textbf{\bm{$R_{cc}$} $\uparrow$} &
\textbf{\bm{$R_{\max}$} $\uparrow$} &
\textbf{\bm{$R_{\min}$} $\uparrow$} &
\textbf{\bm{$R_{\text{sim}}$} $\uparrow$} \\
\midrule

\rowcolor{blockbg}
\multicolumn{7}{l}{\textbf{CL}}\\
\quad + Random
  & 49.0 & -- & -0.312 & 0.999 & -1.384 & -0.255 \\
\quad + QuantAgent
  & \textcolor{bestcolor}{\textbf{57.7}} & \textcolor{bestcolor}{\textbf{+17.8\%}} & \textcolor{bestcolor}{\textbf{-0.195}} & \textcolor{bestcolor}{\textbf{1.181}} & \textcolor{secondbestcolor}{-1.202} & \textcolor{bestcolor}{\textbf{-0.133}} \\
\rowcolor{ourworkbg}
\quad \textbf{+ META (Ours)}
  & \textcolor{secondbestcolor}{56.0} & \textcolor{secondbestcolor}{+6.1\%} & \textcolor{secondbestcolor}{-0.484} & \textcolor{secondbestcolor}{1.024} & \textcolor{bestcolor}{\textbf{-1.187}} & \textcolor{secondbestcolor}{-0.433} \\
\midrule

\rowcolor{blockbg}
\multicolumn{7}{l}{\textbf{ES}}\\
\quad + Random
  & 41.3 & -- & 0.006 & 0.560 & -0.539 & 0.006 \\
\quad + QuantAgent
  & \textcolor{secondbestcolor}{55.0} & \textcolor{secondbestcolor}{+33.2\%} & \textcolor{secondbestcolor}{0.179} & \textcolor{bestcolor}{\textbf{0.613}} & \textcolor{secondbestcolor}{-0.485} & \textcolor{secondbestcolor}{0.179} \\
\rowcolor{ourworkbg}
\quad \textbf{+ META (Ours)}
  & \textcolor{bestcolor}{\textbf{64.0}} & \textcolor{bestcolor}{\textbf{+55.0\%}} & \textcolor{bestcolor}{\textbf{0.278}} & \textcolor{secondbestcolor}{0.584} & \textcolor{bestcolor}{\textbf{-0.431}} & \textcolor{bestcolor}{\textbf{0.250}} \\
\midrule

\rowcolor{blockbg}
\multicolumn{7}{l}{\textbf{NQ}}\\
\quad + Random
  & 41.3 & -- & -0.033 & 0.717 & -0.736 & -0.033 \\
\quad + QuantAgent
  & \textcolor{secondbestcolor}{53.3} & \textcolor{secondbestcolor}{+29.1\%} & \textcolor{secondbestcolor}{0.078} & \textcolor{bestcolor}{\textbf{0.747}} & \textcolor{secondbestcolor}{-0.705} & \textcolor{bestcolor}{\textbf{0.078}} \\
\rowcolor{ourworkbg}
\quad \textbf{+ META (Ours)}
  & \textcolor{bestcolor}{\textbf{62.0}} & \textcolor{bestcolor}{\textbf{+50.1\%}} & \textcolor{bestcolor}{\textbf{0.089}} & \textcolor{secondbestcolor}{0.631} & \textcolor{bestcolor}{\textbf{-0.667}} & \textcolor{secondbestcolor}{-0.029} \\
\midrule

\rowcolor{blockbg}
\multicolumn{7}{l}{\textbf{QQQ}}\\
\quad + Random
  & 39.7 & -- & -0.265 & 0.966 & -1.038 & -0.252 \\
\quad + QuantAgent
  & \textcolor{secondbestcolor}{59.7} & \textcolor{secondbestcolor}{+50.4\%} & \textcolor{secondbestcolor}{0.189} & \textcolor{secondbestcolor}{1.052} & \textcolor{bestcolor}{\textbf{-0.952}} & \textcolor{bestcolor}{\textbf{0.193}} \\
\rowcolor{ourworkbg}
\quad \textbf{+ META (Ours)}
  & \textcolor{bestcolor}{\textbf{61.5}} & \textcolor{bestcolor}{\textbf{+54.9\%}} & \textcolor{bestcolor}{\textbf{0.627}} & \textcolor{bestcolor}{\textbf{1.094}} & \textcolor{secondbestcolor}{-1.102} & \textcolor{secondbestcolor}{-0.029} \\
\midrule

\rowcolor{blockbg}
\multicolumn{7}{l}{\textbf{BTC}}\\
\quad + Random
  & 44.3 & -- & -0.259 & 1.115 & -1.366 & -0.277 \\
\quad + QuantAgent
  & \textcolor{secondbestcolor}{50.7} & \textcolor{secondbestcolor}{+14.5\%} & \textcolor{secondbestcolor}{0.081} & \textcolor{secondbestcolor}{1.232} & \textcolor{secondbestcolor}{-1.249} & \textcolor{bestcolor}{\textbf{0.004}} \\
\rowcolor{ourworkbg}
\quad \textbf{+ META (Ours)}
  & \textcolor{bestcolor}{\textbf{56.0}} & \textcolor{bestcolor}{\textbf{+26.4\%}} & \textcolor{bestcolor}{\textbf{0.120}} & \textcolor{bestcolor}{\textbf{1.296}} & \textcolor{bestcolor}{\textbf{-0.980}} & \textcolor{secondbestcolor}{-0.040} \\
\bottomrule
\end{tabular}
}
\caption{Performance comparison across assets. Accuracy ($\alpha$), relative improvement ($\Delta\alpha$), correlation $R_{cc}$, maximum and minimum returns ($R_{\max}$, $R_{\min}$), and similarity metric ($R_{\text{sim}}$). \textcolor{bestcolor}{Red} indicates better performance, 
\textcolor{secondbestcolor}{Green} indicates worse, and black numbers denote baseline or neutral results.}
\label{tab:results}
\end{table*}

\subsection{Main Results}

Table~\ref{tab:results} presents a comprehensive comparison between \textsc{META},  \textsc{QuantAgent}, and a random trading policy across five assets. The results demonstrate that \textsc{META} consistently outperforms both baselines in directional accuracy and return-based metrics, confirming the effectiveness of its memory-augmented decision process.

In terms of accuracy ($\alpha$), \textsc{META} achieves the highest performance on all assets except Crude Oil (CL), where it performs competitively with \textsc{QuantAgent}. Notably, on S\&P 500 E-mini (ES), accuracy rises from 55.0\% in \textsc{QuantAgent} to 64.0\% with \textsc{META}, corresponding to a relative improvement of +55.0\%. Similar gains are observed on Nasdaq-100 (NQ) and Bitcoin (BTC), where accuracy improves by +50.1\% and +26.4\%, respectively. Even for more stable equity markets such as QQQ, \textsc{META} maintains a marginal but consistent edge over \textsc{QuantAgent} (61.5\% vs.\ 59.7\%).  Beyond directional accuracy, the return-based metrics further corroborate the effectiveness of \textsc{META}. The model consistently achieves the strongest cumulative return correlations ($R_{cc}$) on ES ($0.278$), QQQ ($0.627$), and BTC ($0.120$), indicating that its predicted trade directions align more reliably with realized price movements compared to prior baselines. Moreover, \textsc{META} demonstrates notably improved downside resilience, achieving the least negative drawdowns ($R_{\min}$) across all assets except QQQ, where it remains competitive.



\begin{figure*}[t]
    \centering
    \begin{subfigure}{0.155\textwidth}
        \includegraphics[width=\linewidth]{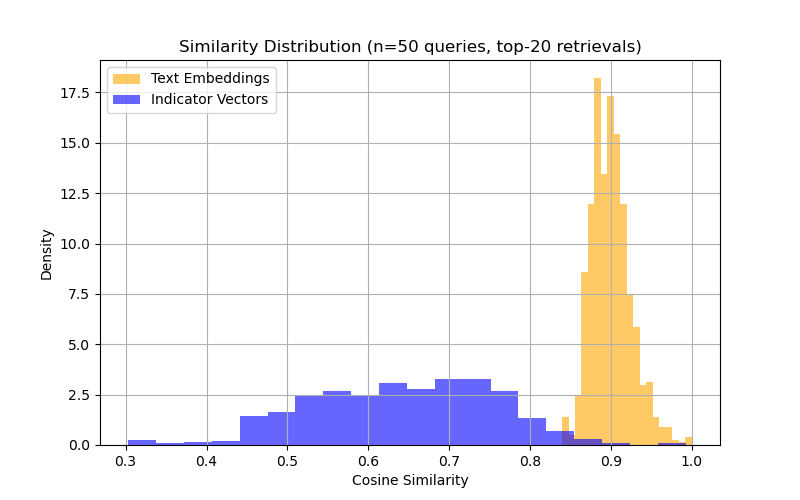}
        \caption{BTC Similarity }
    \end{subfigure}
    \begin{subfigure}{0.155\textwidth}
        \includegraphics[width=\linewidth]{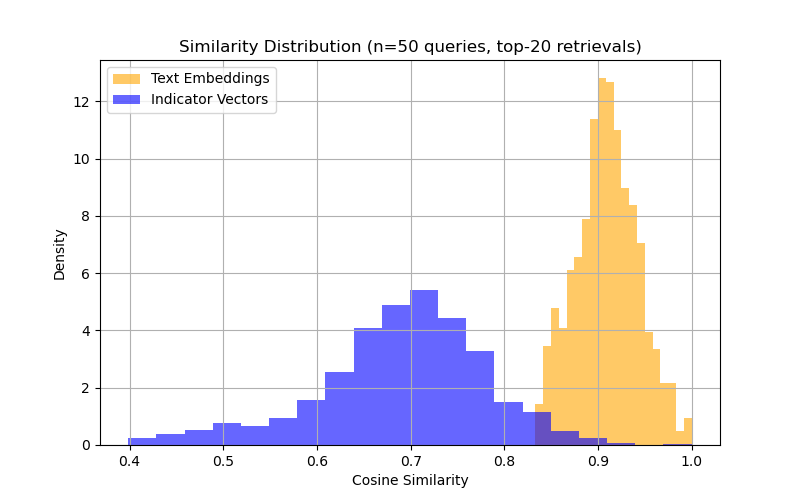}
        \caption{CL Similarity}
    \end{subfigure}
    \begin{subfigure}{0.155\textwidth}
        \includegraphics[width=\linewidth]{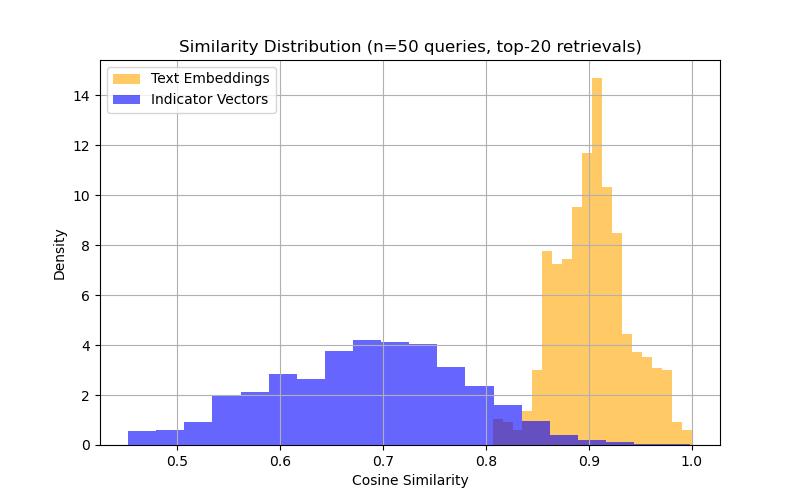}
        \caption{ES Similarity}
    \end{subfigure}
    \begin{subfigure}{0.155\textwidth}
        \includegraphics[width=\linewidth]{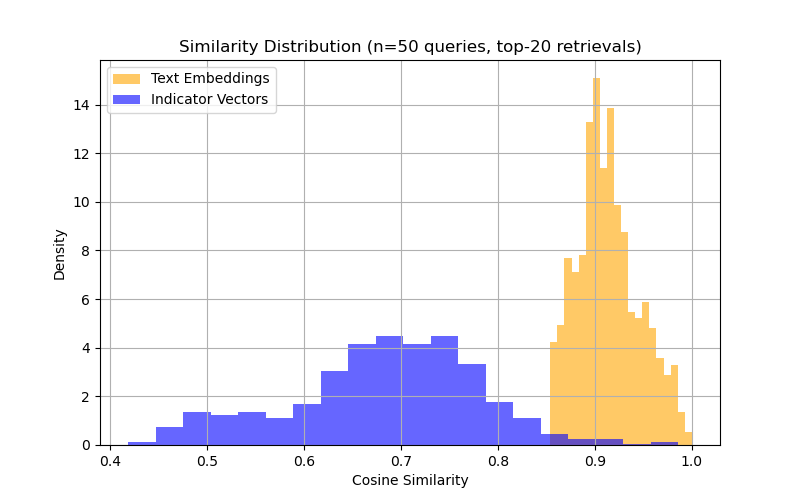}
        \caption{NQ Similarity}
    \end{subfigure}
    \begin{subfigure}{0.155\textwidth}
        \includegraphics[width=\linewidth]{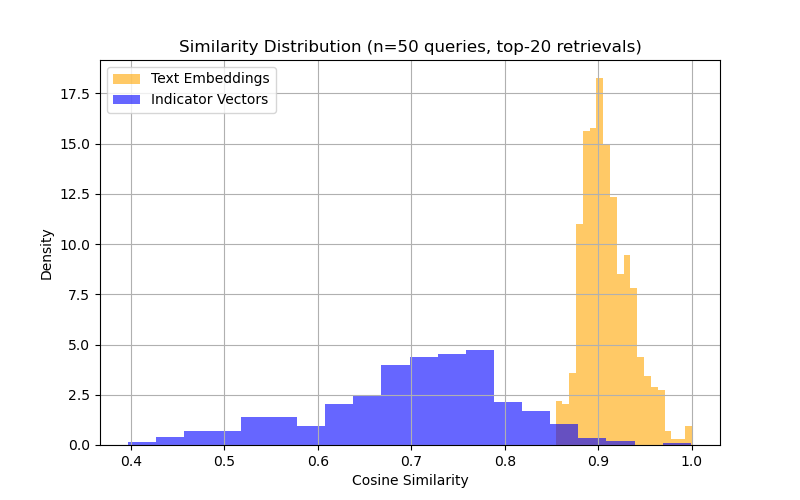}
        \caption{QQQ Similarity}
    \end{subfigure}
    \begin{subfigure}{0.155\textwidth}
        \includegraphics[width=\linewidth]{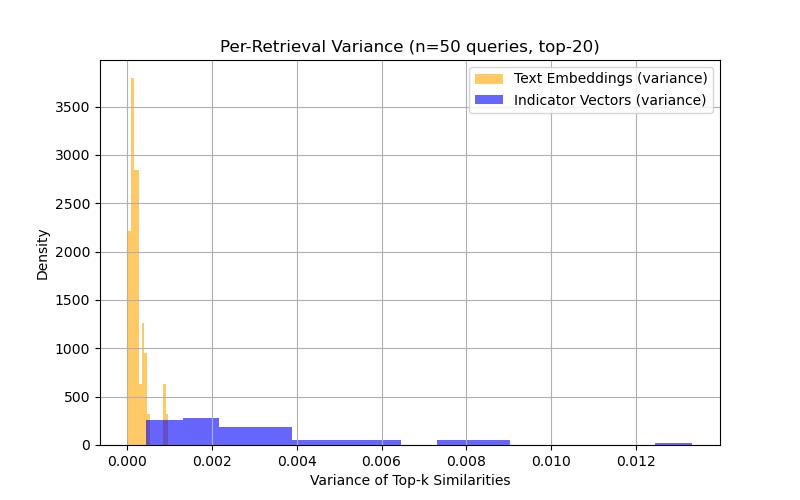}
        \caption{BTC Variance}
    \end{subfigure}
    \begin{subfigure}{0.155\textwidth}
        \includegraphics[width=\linewidth]{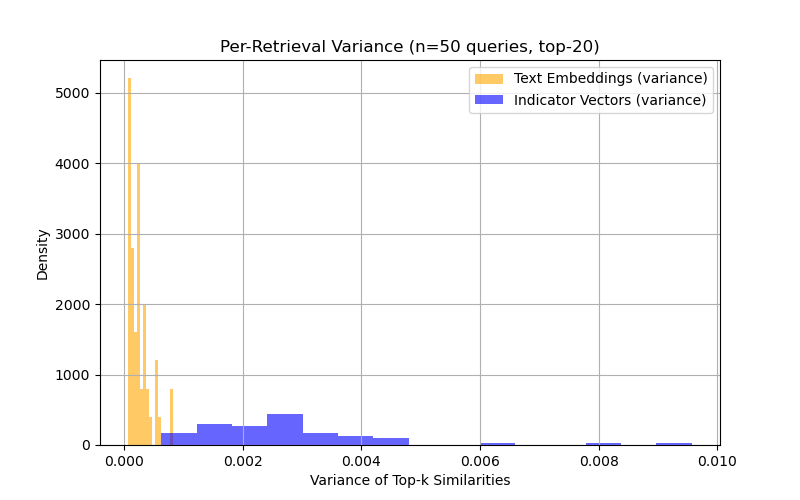}
        \caption{CL Variance}
    \end{subfigure}
    \begin{subfigure}{0.155\textwidth}
        \includegraphics[width=\linewidth]{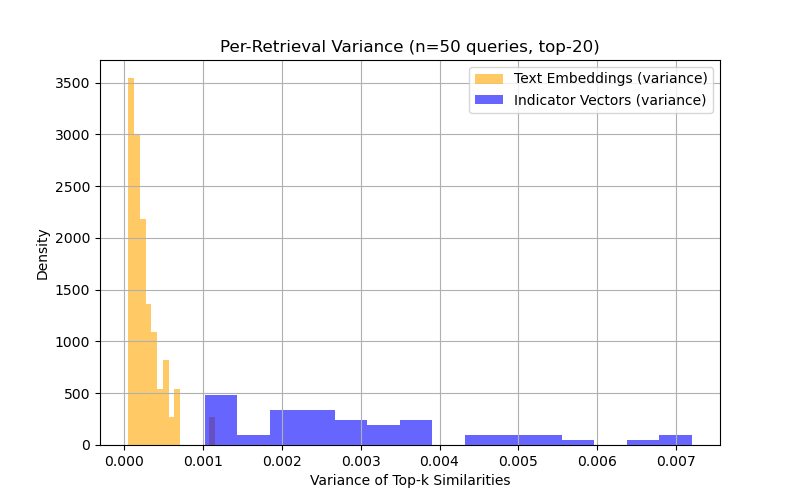}
        \caption{ES Variance}
    \end{subfigure}
    \begin{subfigure}{0.155\textwidth}
        \includegraphics[width=\linewidth]{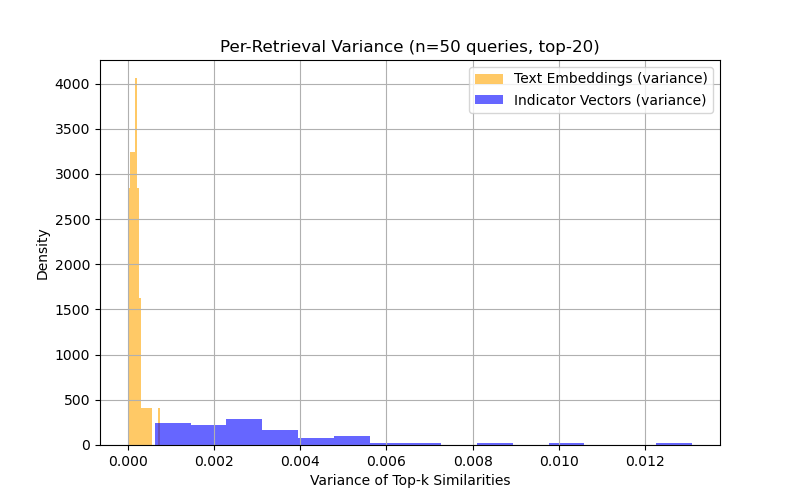}
        \caption{NQ Variance}
    \end{subfigure}
    \begin{subfigure}{0.155\textwidth}
        \includegraphics[width=\linewidth]{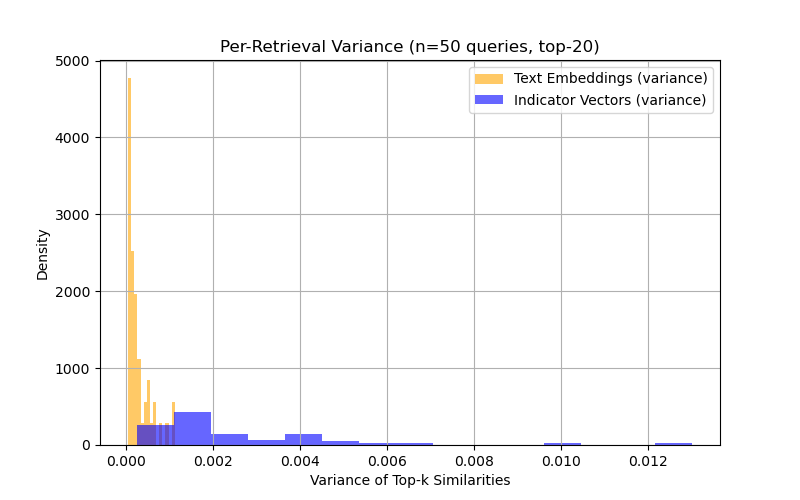}
        \caption{QQQ Variance}
    \end{subfigure}

    \caption{Similarity (top row per asset) and variance (bottom row) distributions for text embeddings vs.\ indicator vectors across all assets.}
    \label{fig:retrieval_grid}
\end{figure*}

\subsection{Text Embeddings vs. Indicator Vectors}

\begin{figure}[h]
    \centering
    \includegraphics[width=0.68\linewidth]{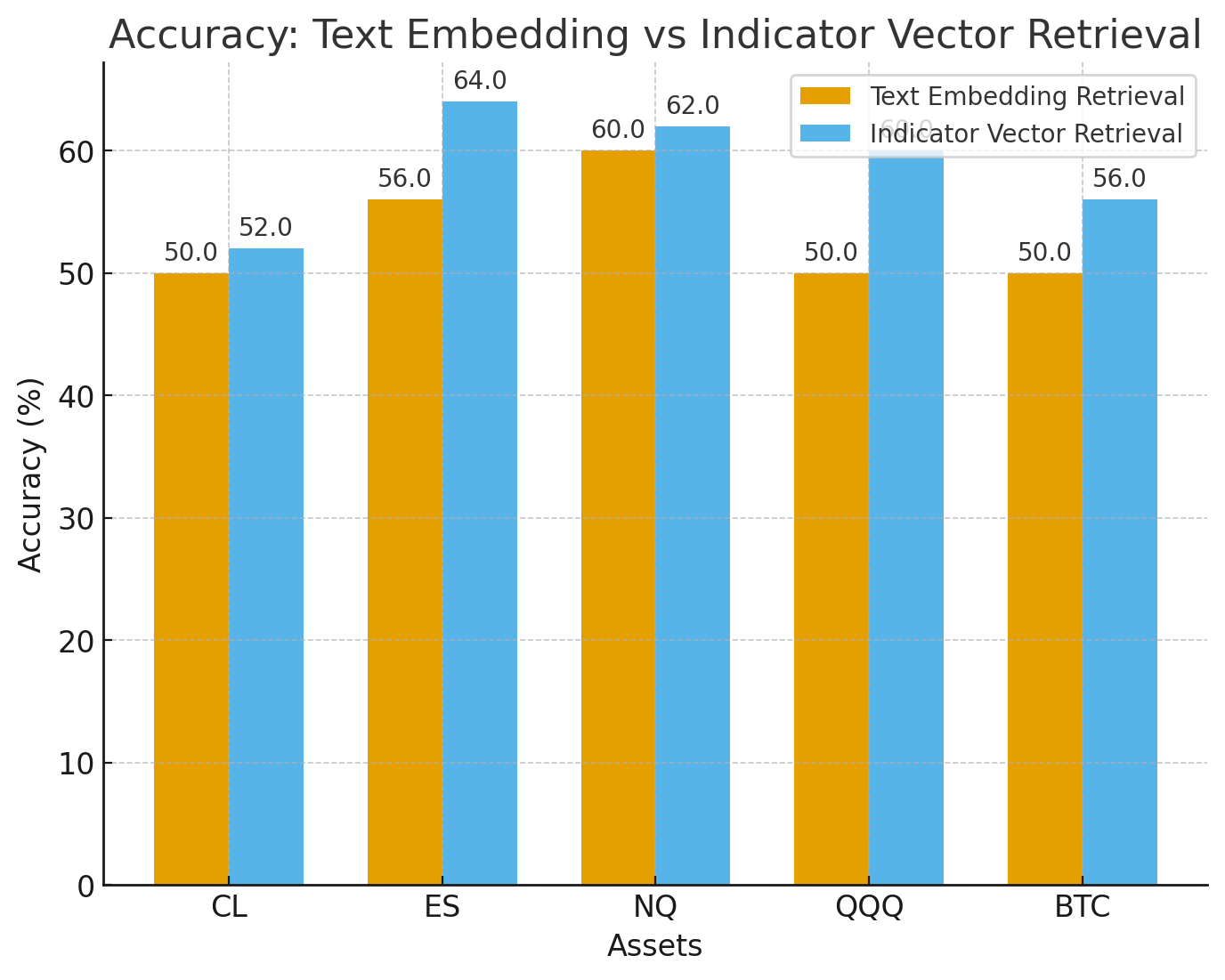}
    \caption{Accuracy comparison of text-embedding vs.\ indicator-vector retrieval.}
    \label{fig:retrieval_acc}
\end{figure}

We compare META’s performance under two retrieval modalities: text embeddings derived from agent reports versus indicator vectors composed of technical features and normalized OHLCV data.As shown in Figure~\ref{fig:retrieval_acc}, indicator-vector retrieval consistently outperforms text-embedding retrieval. On ES futures, accuracy improves from 56.0\% to 64.0\%, and on QQQ from 50.0\% to 60.0\%. Even for more stable markets such as CL and BTC, indicator retrieval yields gains of 2–6 percentage points. Overall, indicator-based retrieval achieves 2–10\% higher accuracy than text embeddings.

\begin{table}[h]
\centering
\renewcommand{\arraystretch}{0.95}
\setlength{\tabcolsep}{8pt}
\small
\begin{tabular}{l|cc|cc}
\toprule
\textbf{Asset} &
\multicolumn{2}{c|}{\textbf{Similarity (Mean)}} &
\multicolumn{2}{c}{\textbf{Variance (Mean)}} \\
\cmidrule(lr){2-3} \cmidrule(lr){4-5}
& \textbf{Text} & \textbf{Indic.} & \textbf{Text} & \textbf{Indic.} \\
\midrule
\rowcolor{blockbg}
BTC  & 0.9002 & 0.6428 & 0.0002 & 0.0031 \\
CL   & 0.9089 & 0.6905 & 0.0002 & 0.0028 \\
\rowcolor{blockbg}
ES   & 0.9036 & 0.6853 & 0.0003 & 0.0030 \\
NQ   & 0.9139 & 0.6826 & 0.0002 & 0.0033 \\
\rowcolor{blockbg}
QQQ  & 0.9095 & 0.7011 & 0.0003 & 0.0026 \\
\bottomrule
\end{tabular}
\caption{Comparison of mean similarity and variance between text embeddings and indicator vectors. Indicator vectors show lower similarity but higher variance, leading to more discriminative retrieval.}
\label{tab:sim_var_means}
\end{table}

Beyond accuracy, the two modalities exhibit distinct retrieval geometries~\cite{zhang2024cross}. Text embeddings produce uniformly high similarity scores (0.90–0.91) with minimal variance ($\approx 0.0002$–$0.0003$), implying that semantically similar reports cluster together regardless of market context—limiting discriminative power. In contrast, indicator vectors yield lower mean similarities (0.64–0.70) but higher variance ($\approx 0.0026$–$0.0033$), indicating sharper separation between relevant and irrelevant precedents. This sharper separation improves retrieval selectivity, which is crucial in high-frequency trading where noisy matches quickly degrade performance. These findings confirm META’s design choice to ground memory retrieval in structured indicator vectors rather than textual embeddings. 
The advantage of compact, structured numeric features over dense semantic embeddings also echoes results in few-shot tabular learning, where LLM-derived structured representations substantially strengthen downstream predictors \cite{yang2026forestllm}.

\begin{figure*}[t]
    \centering
    \begin{subfigure}{0.148\textwidth}
        \includegraphics[width=\linewidth]{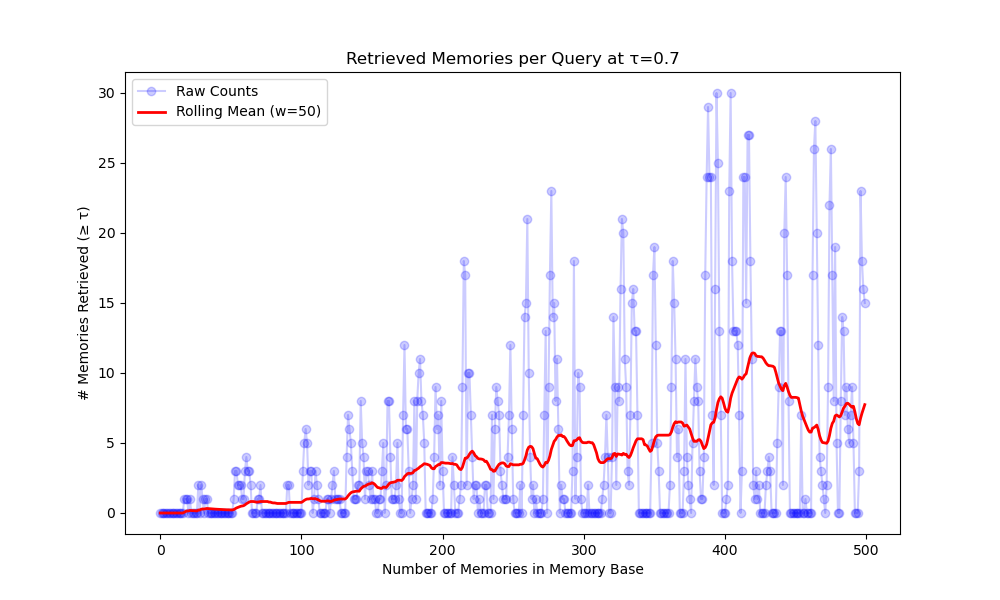}
    \end{subfigure}
    \begin{subfigure}{0.148\textwidth}
        \includegraphics[width=\linewidth]{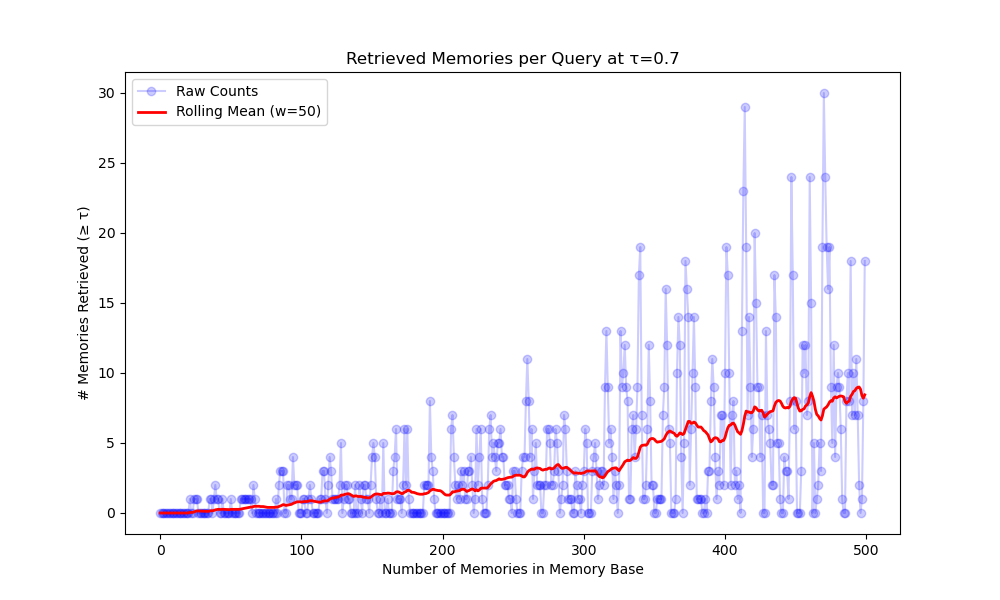}
    \end{subfigure}
    \begin{subfigure}{0.148\textwidth}
        \includegraphics[width=\linewidth]{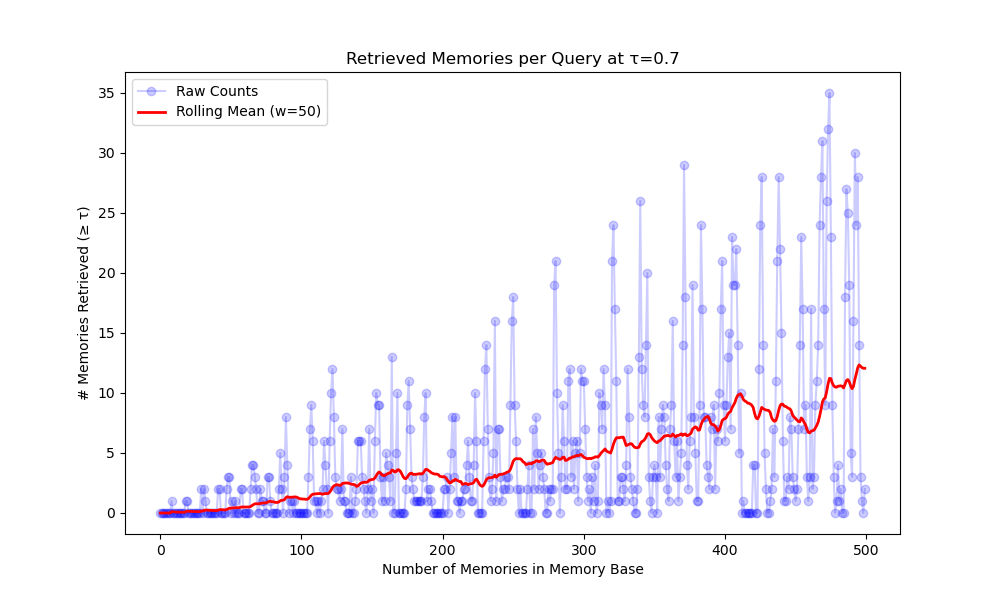}
    \end{subfigure}
    \begin{subfigure}{0.148\textwidth}
        \includegraphics[width=\linewidth]{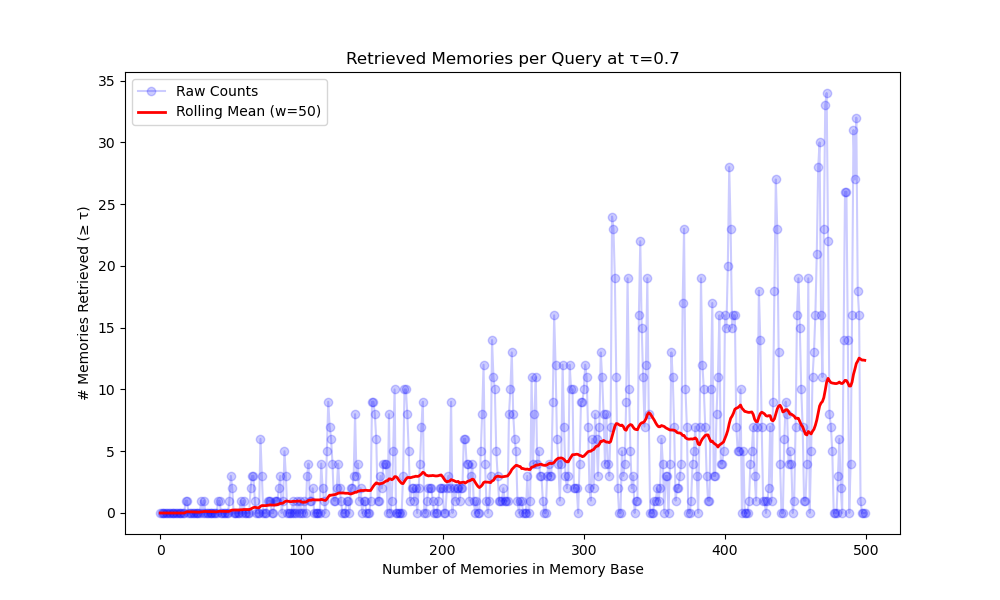}
    \end{subfigure}
    \begin{subfigure}{0.148\textwidth}
        \includegraphics[width=\linewidth]{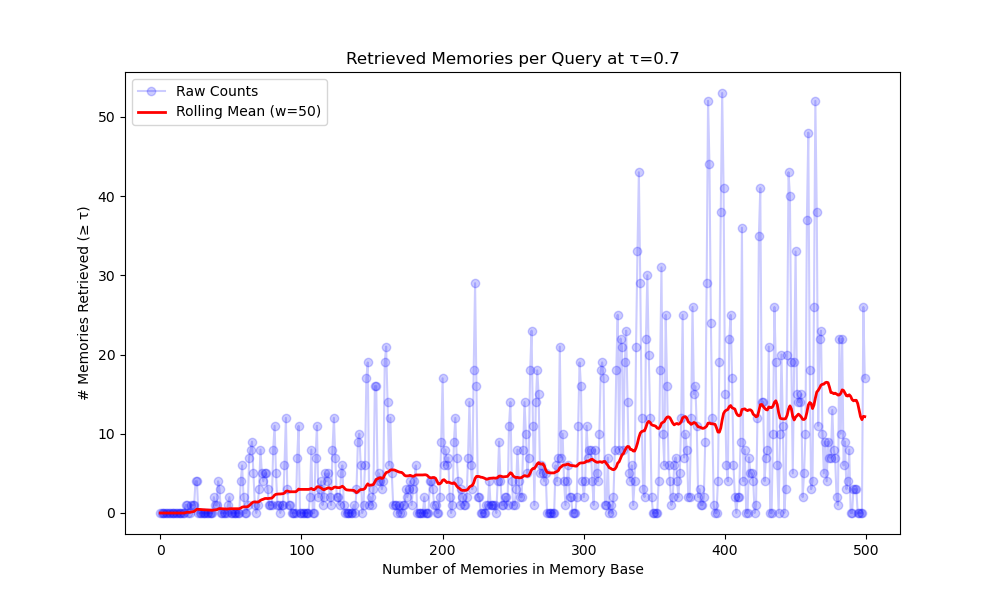}
    \end{subfigure}

    \begin{subfigure}{0.148\textwidth}
        \includegraphics[width=\linewidth]{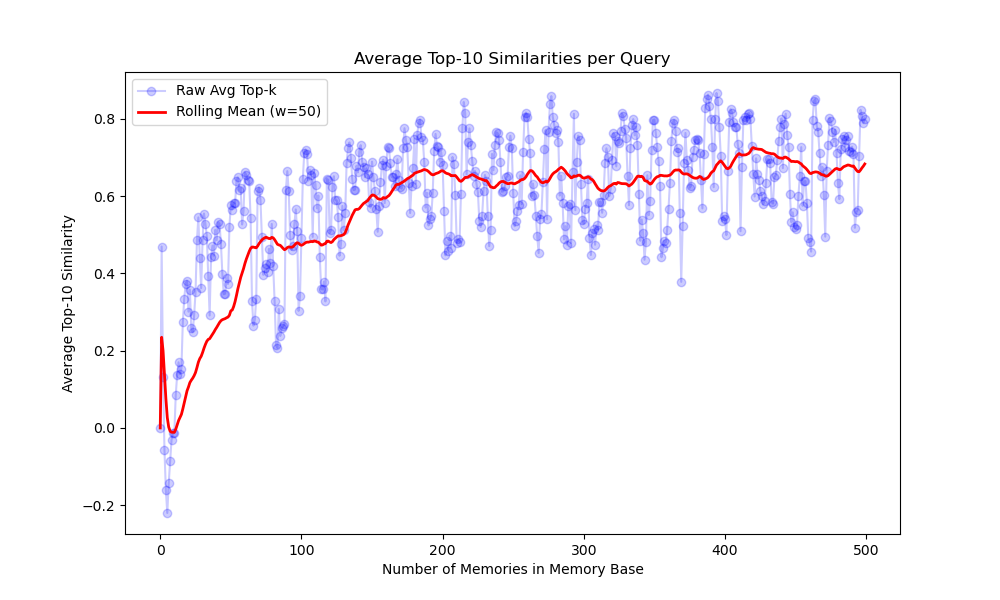}
    \end{subfigure}
    \begin{subfigure}{0.148\textwidth}
        \includegraphics[width=\linewidth]{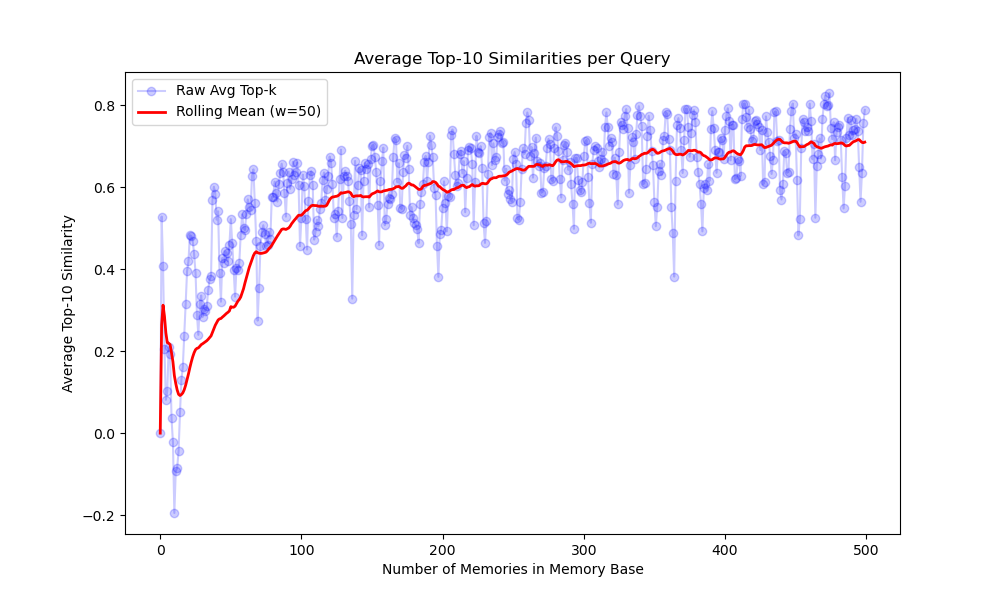}
    \end{subfigure}
    \begin{subfigure}{0.148\textwidth}
        \includegraphics[width=\linewidth]{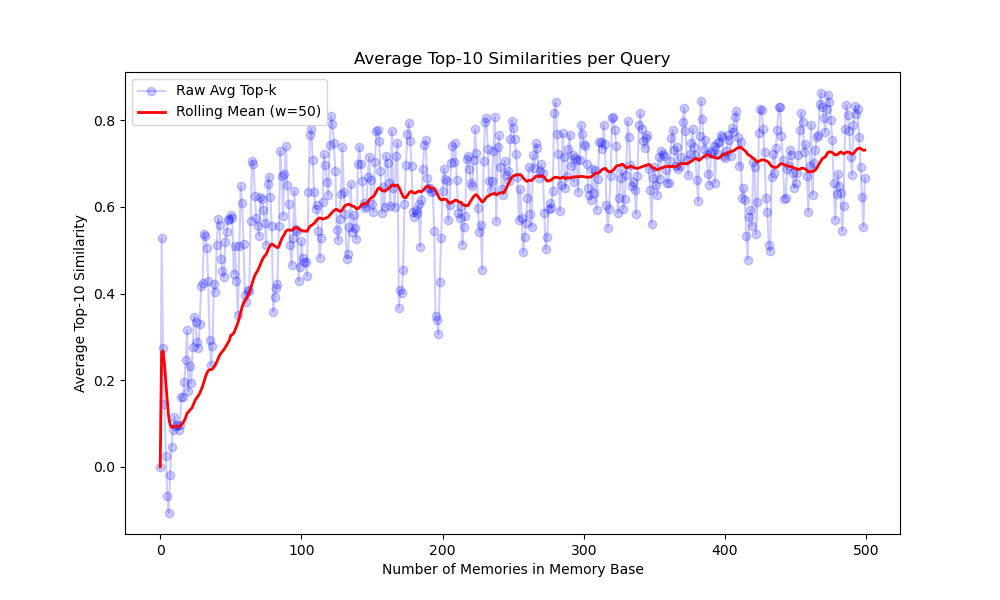}
    \end{subfigure}
    \begin{subfigure}{0.148\textwidth}
        \includegraphics[width=\linewidth]{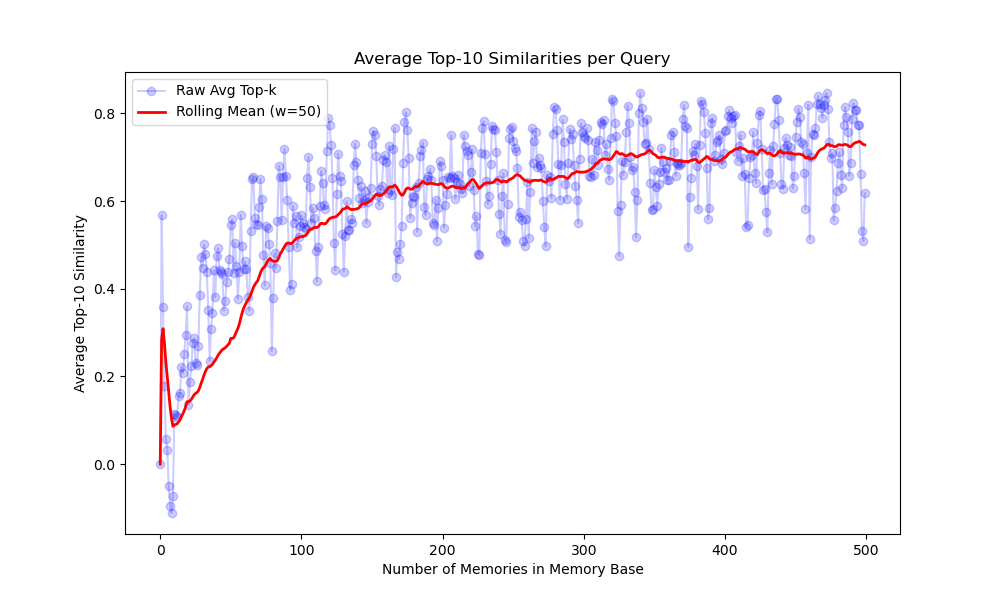}
    \end{subfigure}
    \begin{subfigure}{0.148\textwidth}
        \includegraphics[width=\linewidth]{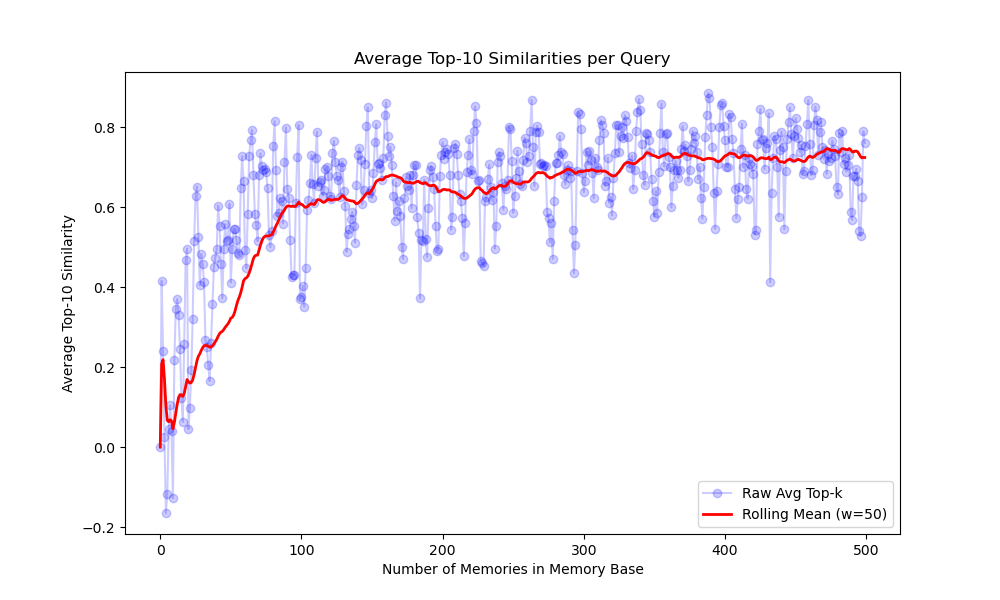}
    \end{subfigure}

    \begin{subfigure}{0.148\textwidth}
        \includegraphics[width=\linewidth]{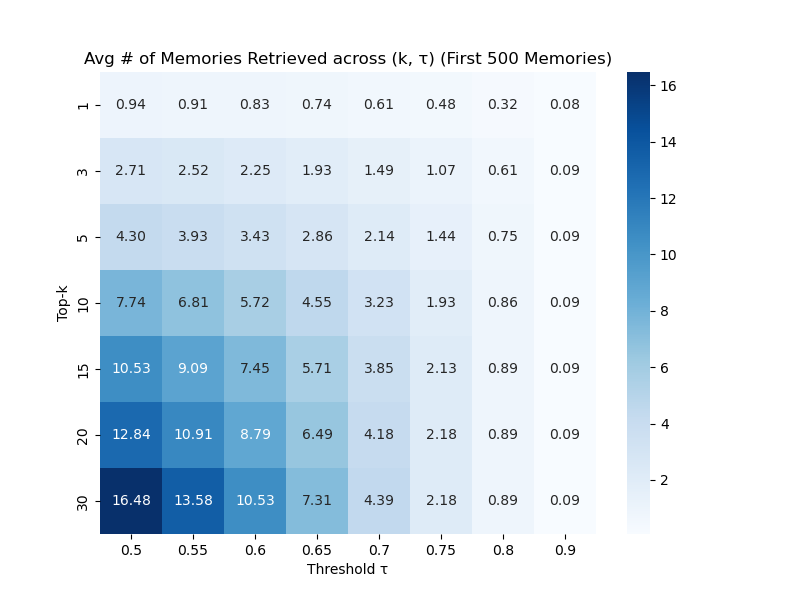}
        \caption{BTC}
    \end{subfigure}
    \begin{subfigure}{0.148\textwidth}
        \includegraphics[width=\linewidth]{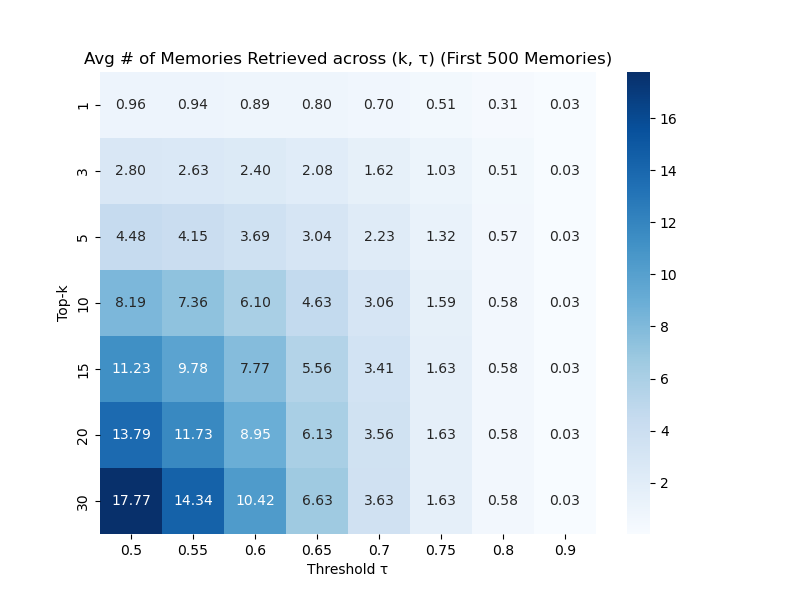}
        \caption{CL}
    \end{subfigure}
    \begin{subfigure}{0.148\textwidth}
        \includegraphics[width=\linewidth]{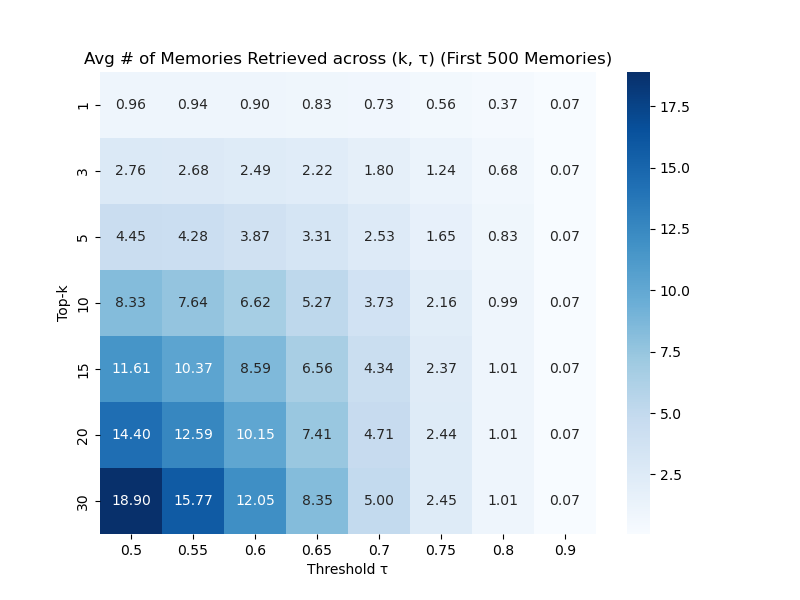}
        \caption{ES}
    \end{subfigure}
    \begin{subfigure}{0.148\textwidth}
        \includegraphics[width=\linewidth]{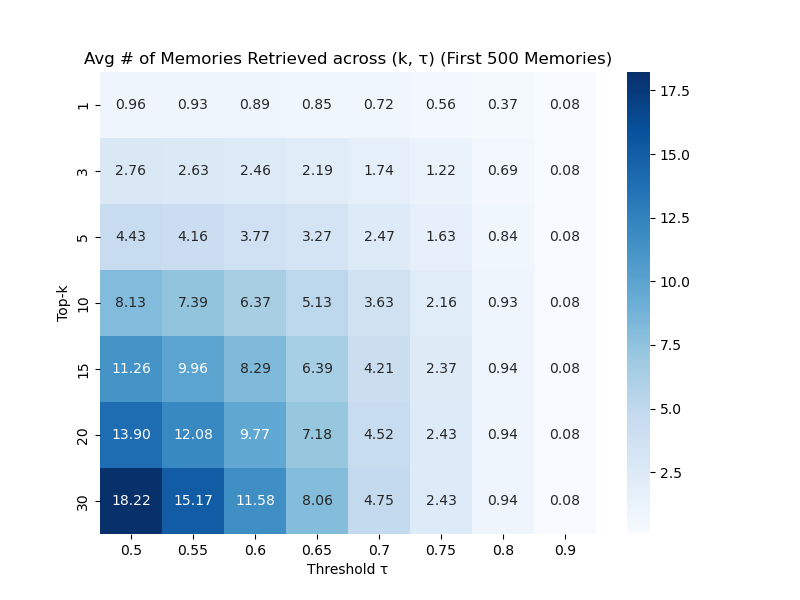}
        \caption{NQ}
    \end{subfigure}
    \begin{subfigure}{0.148\textwidth}
        \includegraphics[width=\linewidth]{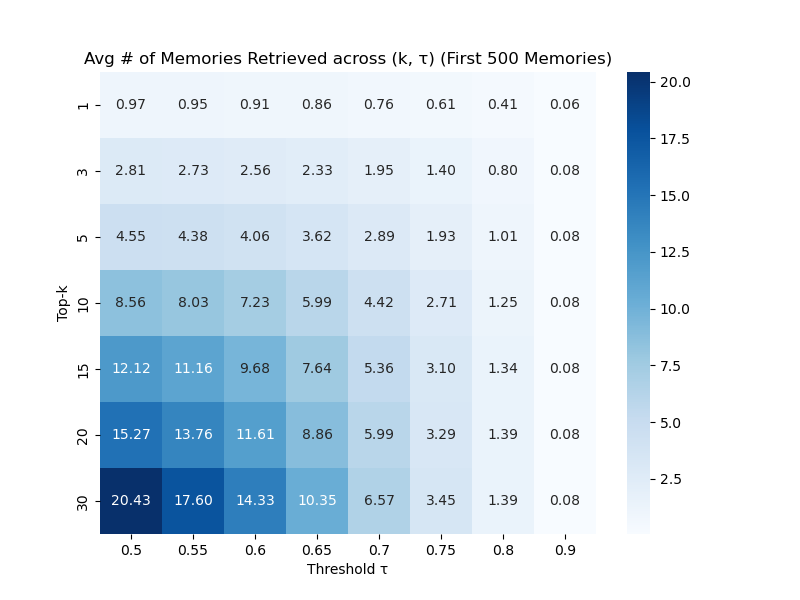}
        \caption{QQQ}
    \end{subfigure}

    \caption{Top-$k$ and threshold retrieval analysis. For each asset: \# memories retrieved ($\tau = 0.7$), average top-$k$ similarities ($k = 10$), and heatmap of retrieval counts over $k$ and $\tau$.}
    \label{fig:topk_threshold_grid}
\end{figure*}

\begin{figure*}[t]
    \centering
    \begin{subfigure}{0.165\textwidth}
        \includegraphics[width=\linewidth]{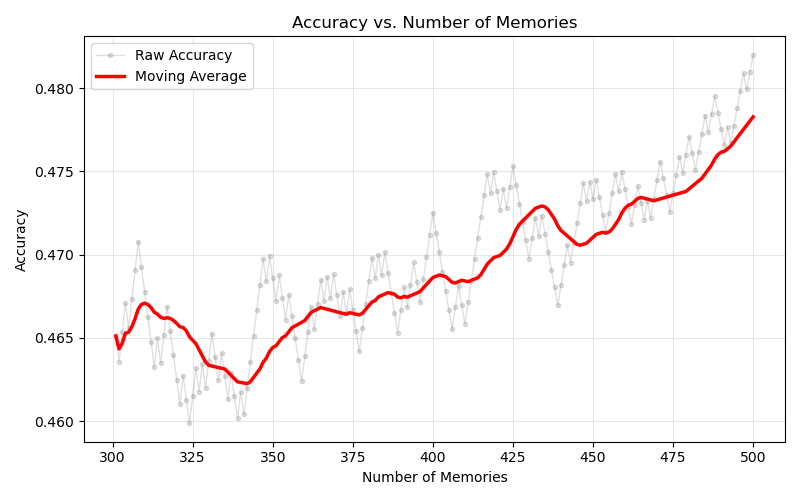}
        \caption{CL}
    \end{subfigure}
    \begin{subfigure}{0.165\textwidth}
        \includegraphics[width=\linewidth]{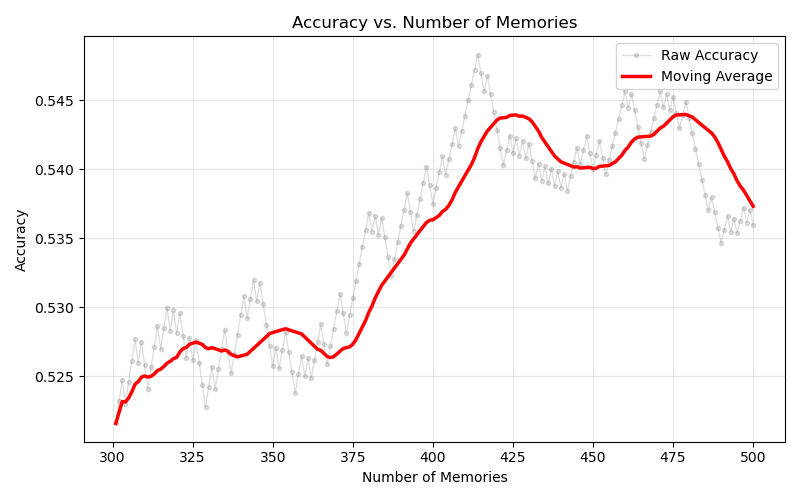}
        \caption{ES}
    \end{subfigure}
    \begin{subfigure}{0.165\textwidth}
        \includegraphics[width=\linewidth]{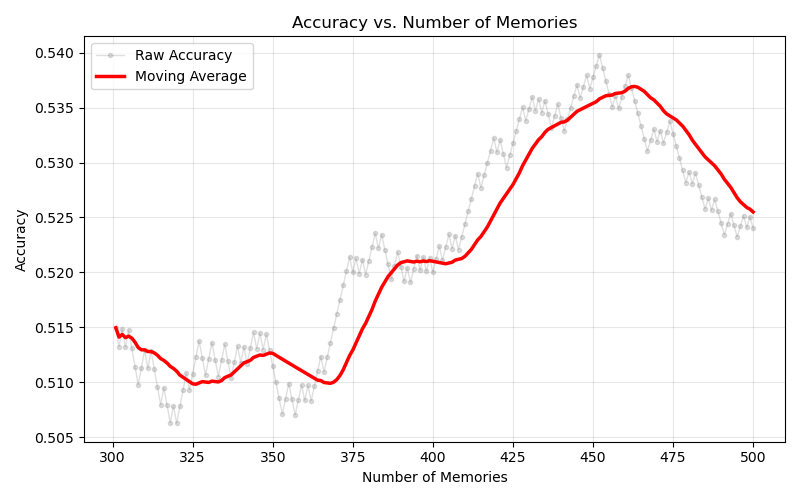}
        \caption{NQ}
    \end{subfigure}
    \begin{subfigure}{0.165\textwidth}
        \includegraphics[width=\linewidth]{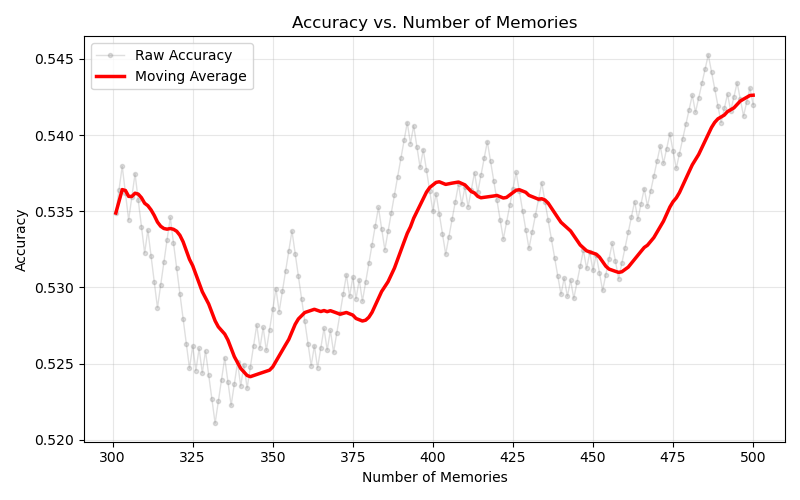}
        \caption{QQQ}
    \end{subfigure}
    \begin{subfigure}{0.165\textwidth}
        \includegraphics[width=\linewidth]{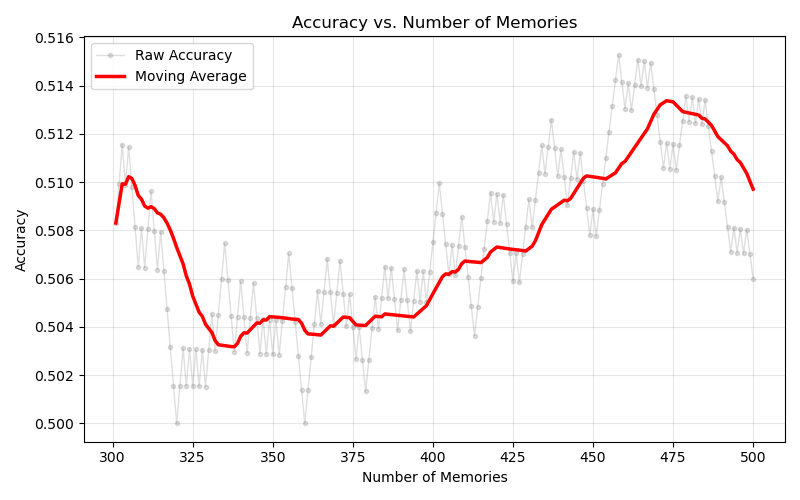}
        \caption{BTC}
    \end{subfigure}
    \caption{Accuracy versus number of memories. Each curve shows the moving average of directional accuracy as META’s memory base expands. The analysis begins at 300 memories, corresponding to the point where episodic retrieval starts to meaningfully influence decision-making.}
    \label{fig:acc_vs_mem}
\end{figure*}

\subsection{Memory Retrieval Guidance }

We analyze how the top-$k$ and similarity threshold ($\tau$) hyperparameters control retrieval quality and efficiency, validating META’s choice of combining both mechanisms for stable and adaptive recall.

\paragraph{Retrieved counts at fixed threshold.}  
At a fixed threshold (e.g., $\tau=0.7$), the number of retrieved memories per query (Figure~\ref{fig:topk_threshold_grid}, top) increases steadily as the memory base grows. Although individual counts fluctuate, the rolling average rises over time, showing that thresholding functions as a quality filter—restricting retrieval early when memory is sparse, and gradually broadening it as more relevant experiences accumulate.

\paragraph{Average top-$k$ similarity at fixed $k$.}  
With $k$ fixed (e.g., $k=10$), the average similarity of retrieved memories (Figure~\ref{fig:topk_threshold_grid}, middle) improves as the memory expands, stabilizing at a high level. This demonstrates that top-$k$ selection maintains focus on the most relevant precedents while preventing uncontrolled growth in reference size.

\paragraph{Joint effect of $k$ and $\tau$.}  
Heatmaps of average retrieval counts across $k$--$\tau$ combinations (Figure~\ref{fig:topk_threshold_grid}, bottom) reveal their complementary dynamics. At low $\tau$, increasing $k$ rapidly enlarges the retrieved set, while very high $\tau$ collapses retrieval to near zero. Balanced behavior occurs around moderate settings ($\tau=0.6$--0.7, $k=5$--15), yielding both sufficient coverage and high reliability.

\begin{table}[t]
\centering
\renewcommand{\arraystretch}{0.9}
\setlength{\tabcolsep}{4pt} 
\footnotesize
\begin{tabular}{l|cc|cc}
\toprule
\textbf{Asset} &
\textbf{Text Acc.} &
\textbf{Indic. Acc.} &
\textbf{Text Time} &
\textbf{Indic. Time} \\
\midrule
\rowcolor{blockbg}
CL   & 50.0 & \textcolor{bestcolor}{\textbf{52.0}} & 3.11 & 0.00610 \\
ES   & 56.0 & \textcolor{bestcolor}{\textbf{64.0}} & 3.18 & 0.00580 \\
\rowcolor{blockbg}
NQ   & 60.0 & \textcolor{bestcolor}{\textbf{62.0}} & 3.15 & 0.00595 \\
QQQ  & 50.0 & \textcolor{bestcolor}{\textbf{61.0}} & 3.12 & 0.00620 \\
\rowcolor{blockbg}
BTC  & 50.0 & \textcolor{bestcolor}{\textbf{56.0}} & 3.17 & 0.00575 \\
\bottomrule
\end{tabular}
\caption{Accuracy (\%) and retrieval latency (s) comparison between text-embedding and indicator-vector memory retrieval.}
\label{tab:latency_comparison}
\end{table}

\subsection{Latency-Aware Retrieval Analysis}
\label{subsec:latency_analysis}

Latency is a critical constraint in high-frequency trading, META therefore adopts two latency-aware design choices relative to prior LLM-based trading agents: (i) all Signal Agents operate in parallel, avoiding stacking agentic reasoning process; and (ii) episodic memory is retrieved using cosine similarity in indicator-vector space rather than semantic text embeddings. Table~\ref{tab:latency_comparison} summarizes the accuracy--latency tradeoff across five assets. Indicator-vector retrieval improves directional accuracy by 2--11 percentage points while reducing retrieval latency from approximately 3.1 seconds to 0.006 seconds per query. This corresponds to an average \textbf{~500$\times$ speedup} in memory access across all assets. The results confirm that indicator vectors enable fast, domain-aligned retrieval without sacrificing accuracy, allowing episodic memory to be integrated directly into the trading loop.

\subsection{Learning Dynamics with Memory Accumulation}

We examine how META’s performance evolves as its memory base grows. Figure~\ref{fig:acc_vs_mem} shows the moving-average directional accuracy. The analysis starts at 300 memories, since the initial 0--300 trades rely primarily on perception and reasoning modules rather than episodic recall; early accuracy variations thus reflect exploratory behavior rather than learned adaptation. Beyond this phase, all assets exhibit a clear upward trend in accuracy, with a transient volatility window around 300--350 trades as retrieval stabilizes and the memory base diversifies. As memory density increases further, accuracy steadily improves and variance declines, confirming that broader episodic coverage enhances contextual reasoning and decision precision. Notably, higher-volatility assets such as BTC and NQ display sharper accuracy gains, highlighting the value of episodic retrieval in turbulent markets.

\begin{figure}[h]
    \centering
    \includegraphics[width=0.78\linewidth]{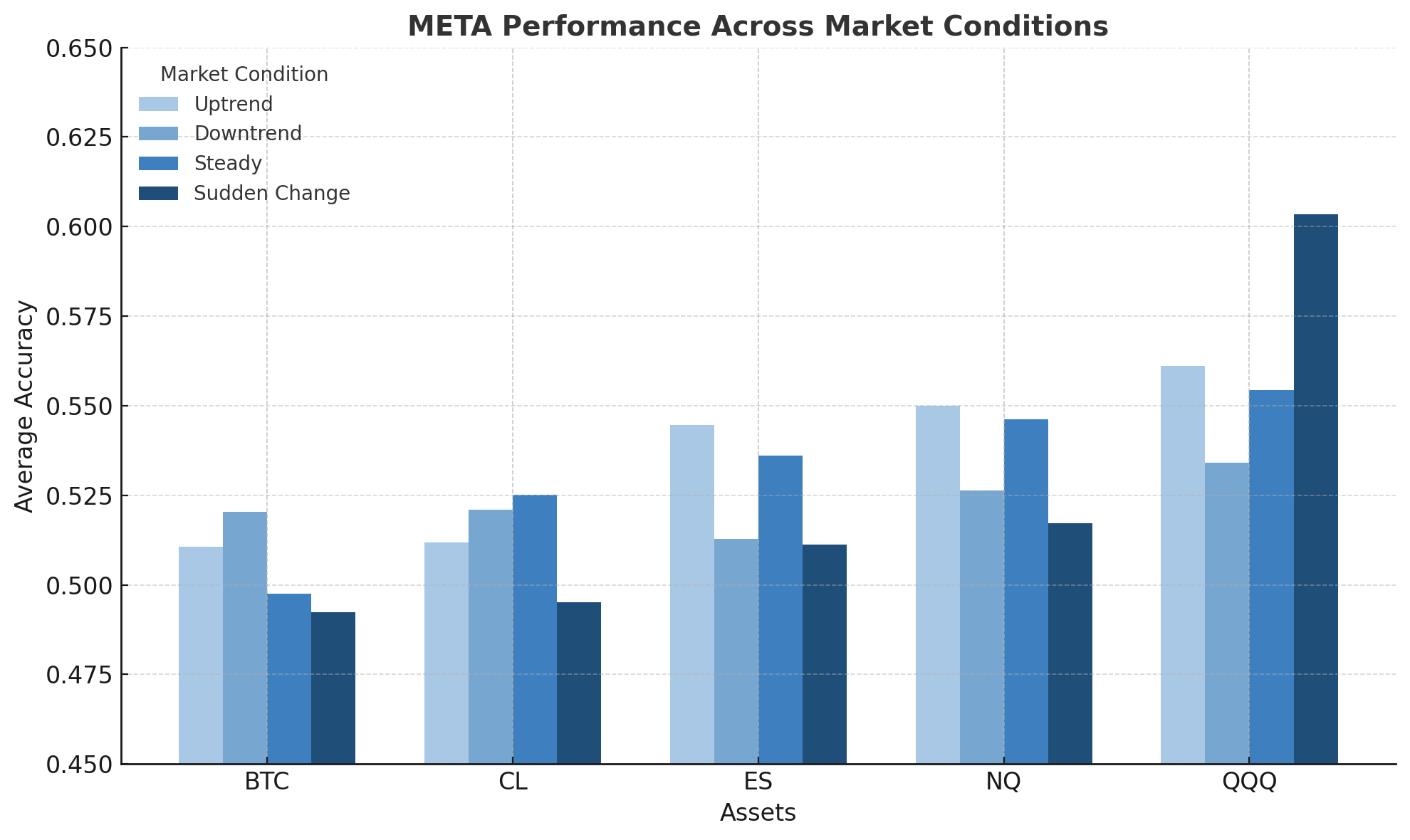}
    \caption{Performance of \textsc{META} under different market conditions across assets. }
    \label{fig:robustness_bar}
\end{figure}

\subsection{Robustness Under Varying Market Conditions}

We evaluate \textsc{META}’s stability across four representative market regimes—\textit{uptrend}, \textit{downtrend}, \textit{steady}, and \textit{sudden change}—to assess its adaptability under dynamic trading environments. As shown in Figure~\ref{fig:robustness_bar}, performance remains highly consistent across regimes, with accuracy variations confined to within 4–6\% across all assets.

Across the five assets, \textsc{META} achieves its greatest stability in steady markets (average accuracy above 0.53) by exploiting recurring structural patterns through memory retrieval, while comparable accuracy in \textit{uptrend} and \textit{downtrend} phases indicates effective generalization without directional bias. The framework is also resilient during sudden price shocks (e.g., 0.60 accuracy on QQQ), as its retrieval-guided decisions draw on structurally similar high-volatility episodes rather than short-term fluctuations.

\begin{table}[!t]
\centering
\renewcommand{\arraystretch}{1} 
\setlength{\tabcolsep}{6pt}        
\small
\begin{tabular}{l|l|cc}
\toprule
\textbf{Asset} & \textbf{Method} & \textbf{Accuracy $\alpha$ (\% ↑)} & \textbf{$\Delta\alpha$ (↑)} \\
\midrule

\rowcolor{blockbg}
CL  & w/o Mem & 52.5 & -- \\
    & \textbf{w/ Mem} & \textcolor{bestcolor}{\textbf{56.0}} & \textcolor{bestcolor}{\textbf{+3.5}} \\
\midrule

ES  & w/o Mem & 50.0 & -- \\
\rowcolor{blockbg}
    & \textbf{w/ Mem} & \textcolor{bestcolor}{\textbf{64.0}} & \textcolor{bestcolor}{\textbf{+14.0}} \\
\midrule

NQ  & w/o Mem & 40.0 & -- \\
    & \textbf{w/ Mem} & \textcolor{bestcolor}{\textbf{62.0}} & \textcolor{bestcolor}{\textbf{+22.0}} \\
\midrule

\rowcolor{blockbg}
QQQ & w/o Mem & 55.0 & -- \\
    & \textbf{w/ Mem} & \textcolor{bestcolor}{\textbf{61.5}} & \textcolor{bestcolor}{\textbf{+6.5}} \\
\midrule

BTC & w/o Mem & 48.0 & -- \\
\rowcolor{blockbg}
    & \textbf{w/ Mem} & \textcolor{bestcolor}{\textbf{56.0}} & \textcolor{bestcolor}{\textbf{+8.0}} \\

\bottomrule
\end{tabular}
\caption{Ablation study of memory system. }

\label{tab:ablation_memory}
\end{table}

\subsection{Memory System Ablation}

We perform an ablation study to isolate the impact of the memory module by comparing \textsc{META}’s performance with and without memory across five assets (Table~\ref{tab:ablation_memory}). Across all settings, incorporating memory yields consistent and substantial gains in directional accuracy.

For high-volatility assets such as ES and NQ, the improvements are especially pronounced: accuracy increases from 50.0\% to 64.0\% (+14.0 points) on ES and from 40.0\% to 62.0\% (+22.0 points) on NQ. These results highlight the value of episodic retrieval in distinguishing subtle recurring patterns that short-term indicators alone fail to capture. Steadier assets also benefit, with CL improving from 52.5\% to 56.0\% (+3.5 points) and QQQ from 55.0\% to 60.0\% (+5.0 points). Even BTC, despite its noisy structure, shows an +8.0 point gain.

Overall, these consistent gains across volatile and stable markets confirm that the memory mechanism enhances predictive robustness and contextual reasoning.

\subsection{Indicator Contributions}

To analyze the internal dynamics of the \textsc{META} framework, we perform an ablation study assessing the contribution of each technical indicator to overall trading performance. Each indicator—\textbf{MACD}, \textbf{AVWAP}, \textbf{Trend}, \textbf{RSI}, \textbf{Heiken Ashi}, \textbf{Stochastic}, and \textbf{SMA}—is removed individually, and the resulting change in directional accuracy quantifies its marginal impact on decision quality.

\begin{figure}[t]
    \centering
    \includegraphics[width=0.72\linewidth]{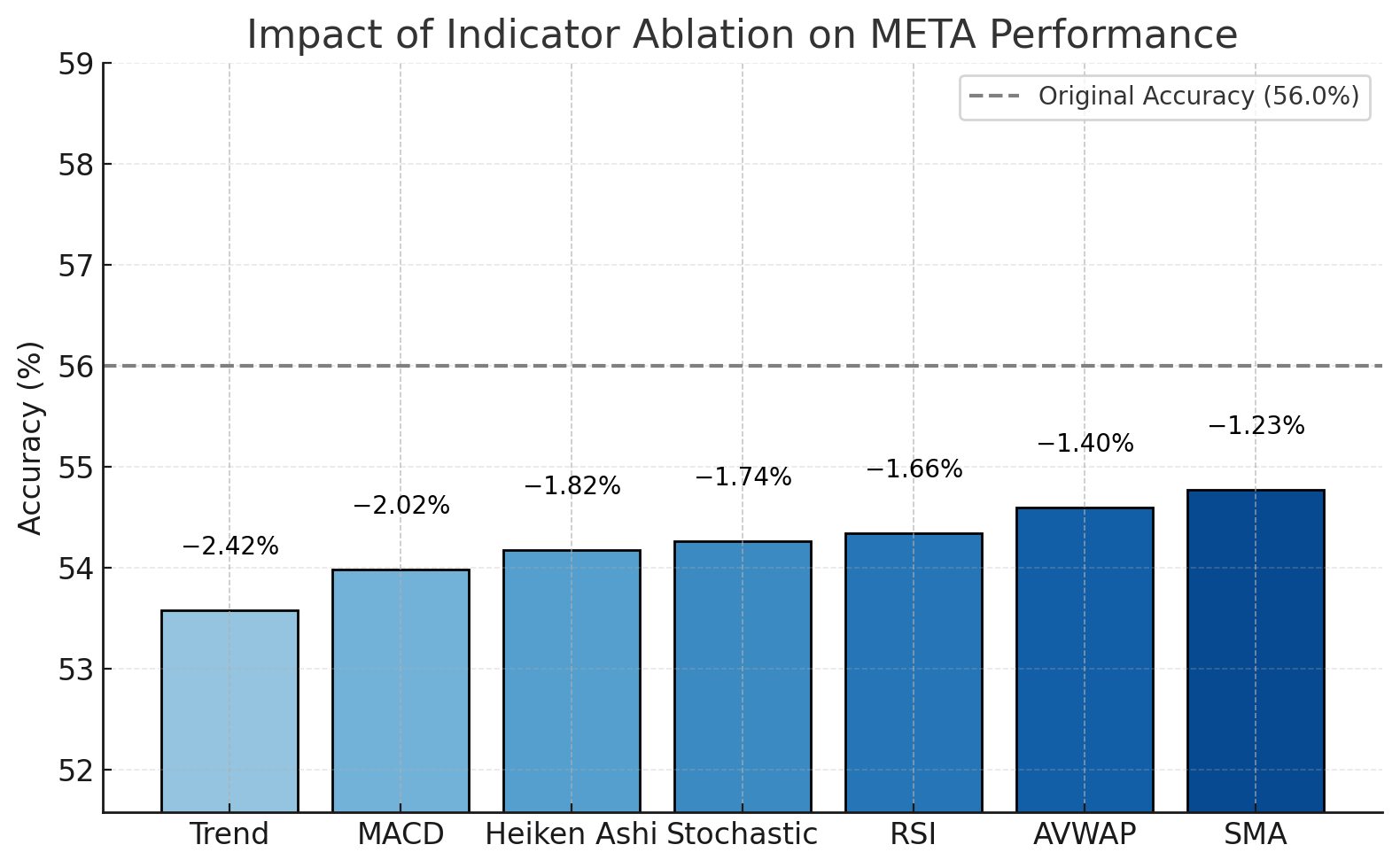}
    \caption{Impact of removing each technical indicator on META’s directional accuracy. Larger decreases indicate greater importance of that indicator to overall system stability and predictive accuracy.}
    \label{fig:indicator_ablation}
\end{figure}

As shown in Figure~\ref{fig:indicator_ablation}, removing any single indicator reduces accuracy, confirming that all components contribute meaningfully to META’s perception of market dynamics. \textbf{Trend} and \textbf{MACD} yield the largest declines ($-2.42\%$ and $-2.02\%$), reflecting their central role in capturing mid- to long-term momentum; \textbf{Heiken Ashi} and \textbf{Stochastic} produce moderate drops ($-1.82\%$ and $-1.74\%$); and \textbf{AVWAP} and \textbf{SMA} show smaller effects ($-1.40\%$ and $-1.23\%$), suggesting partial redundancy with other trend-based indicators.

\section{Conclusion}

We present \textsc{META}, a memory-augmented trading framework that integrates multi-agent market perception with reflective reasoning. By combining specialized Indicator Agents, a decision fusion mechanism, and a dynamic Memory Operator, \textsc{META} enables adaptive and interpretable trading decisions in a zero-shot setting. Extensive experiments across diverse assets show that \textsc{META} consistently outperforms baselines, validating the effectiveness of memory-driven contextual recall for high-frequency decision making. 

\section*{Limitations}

While META demonstrates notable improvements in short-horizon trading accuracy and adaptability, several limitations remain.  
The current framework relies on handcrafted technical indicators for perceptual grounding, which may restrict scalability to more diverse market structures or alternative data modalities.
Although the episodic memory improves contextual reasoning, its similarity-based retrieval remains static; future work could explore adaptive or learned retrieval metrics that evolve with market regimes.  
In addition, our evaluation focuses on retrospective simulation under historical data rather than live market execution, and thus omits slippage, transaction costs, and real-time latency constraints.
Addressing these limitations would move META closer to a fully autonomous, production-ready cognitive trading agent.






\bibliography{custom.bib}

\clearpage

\appendix
\section{Prompts}
\label{sec:prompt}

\subsection*{A.1 MACD Agent}

To instantiate the \textbf{MACDAgent}, we construct a prompt that guides the agent to identify short-term momentum shifts using the \textbf{Moving Average Convergence Divergence (MACD)} indicator. 
Operating under high-frequency trading (HFT) constraints, the agent follows a two-step reasoning process: first generating the MACD chart via tool invocation, then analyzing key momentum patterns such as crossovers, histogram variation, and divergence before issuing a trading signal.

\begin{tcolorbox}[promptbox]
You are a MACD (Moving Average Convergence Divergence) analysis assistant operating in a high-frequency trading context.\\
You must first call the tool \texttt{generate\_macd\_image} using the provided \texttt{kline\_data}.\\
Once the MACD chart is generated, analyze the image for the following key signals:

\begin{enumerate}
    \item \textbf{Signal Line Crossovers}: MACD line crossing above/below the signal line
    \item \textbf{Zero Line Crossovers}: MACD line crossing above/below the zero line
    \item \textbf{Histogram Analysis}: Changes in histogram direction and magnitude
    \item \textbf{Divergences}: Price vs.\ MACD divergences (bullish/bearish)
    \item \textbf{Momentum Strength}: Distance from zero line and slope steepness
\end{enumerate}

Only after generating and analyzing the MACD image should you make a prediction about momentum direction and trading signals.\\
Do not make any predictions before generating and analyzing the image.
\end{tcolorbox}

\textbf{Prompt for MACDAgent in our multi-agent LLM framework.}
The agent receives OHLC data as input and first invokes \texttt{generate\_macd\_image} to obtain a visual depiction of market momentum. 
This prompt enforces a tool-first reasoning loop, ensuring predictions are grounded on the generated chart rather than raw prices.

\begin{tcolorbox}[promptbox]
This is a ---time\_frame--- MACD chart generated from recent OHLC market data. The chart includes: \\
\textbf{MACD Line (blue)} – difference between 12-period and 26-period EMAs;\\
\textbf{Signal Line (red)} – 9-period EMA of MACD line;\\
\textbf{Histogram (bars)} – difference between MACD and Signal lines;\\
\textbf{Zero Line} – reference for bullish/bearish momentum.

Analyze the chart with respect to the following aspects:

\begin{enumerate}
    \item \textbf{Signal Line Crossovers:} Identify whether the MACD line crosses above (bullish) or below (bearish) the signal line and its relation to the zero line.
    \item \textbf{Zero Line Crossovers:} Determine overall momentum and recent transitions across the zero baseline.
    \item \textbf{Histogram Dynamics:} Evaluate bar expansion or contraction to gauge momentum strength.
    \item \textbf{Divergence Patterns:} Compare price action with MACD behavior for bullish or bearish divergences.
    \item \textbf{Momentum Strength:} Assess slope direction and distance from zero to estimate trend intensity.
\end{enumerate}

Provide:
\begin{itemize}
    \item \textbf{Primary Signal:} Buy, Sell, or Hold
    \item \textbf{Momentum Direction:} Bullish, Bearish, or Neutral
    \item \textbf{Confidence Level:} High, Medium, or Low
    \item \textbf{Supporting Evidence:} Observed MACD behavior
    \item \textbf{Risk Considerations:} Potential false or conflicting signals
\end{itemize}

Support your prediction with reasoning and observed evidence.
\end{tcolorbox}

\textbf{Graph-based prompt for MACDAgent in our multi-agent LLM framework.}
Given a generated MACD chart, the agent examines crossovers, histogram movements, and divergence to infer directional bias and confidence. 
This visual grounding enables consistent momentum interpretation across varying market conditions.

\subsection*{A.2 AVWAP Agent}

To instantiate the \textbf{AVWAPAgent}, we construct a prompt that guides the agent to analyze price behavior relative to the \textbf{Anchored Volume Weighted Average Price (AVWAP)}. 
Operating under high-frequency trading (HFT) constraints, the agent combines tool-based chart generation with structured reasoning to detect support/resistance reactions, mean-reversion signals, and breakout dynamics around volume-weighted fair value.

\begin{tcolorbox}[promptbox]
You are an Anchored VWAP (Volume Weighted Average Price) analysis assistant operating in a high-frequency trading context.\\
You must first call the tool \texttt{generate\_anchored\_vwap\_image} using the provided \texttt{kline\_data}.\\
Once the Anchored VWAP chart is generated, analyze the image for the following key patterns:

\begin{enumerate}
    \item \textbf{Support/Resistance Levels}: How price reacts when touching the AVWAP line.
    \item \textbf{Mean Reversion Signals}: Extreme deviations from AVWAP suggesting pullbacks.
    \item \textbf{Breakout Patterns}: Decisive moves above/below AVWAP with volume confirmation.
    \item \textbf{AVWAP Slope Analysis}: Direction and steepness indicating market bias.
    \item \textbf{Volume Confirmation}: Volume spikes during AVWAP interactions.
    \item \textbf{Distance Analysis}: Magnitude of price deviation from fair value anchor.
\end{enumerate}

Only after generating and analyzing the Anchored VWAP image should you make predictions about price direction and trading opportunities.\\
Do not make any predictions before generating and analyzing the image.
\end{tcolorbox}

\textbf{Prompt for AVWAPAgent in our multi-agent LLM framework.}
The agent first generates an Anchored VWAP chart from OHLC data, enforcing a visual-first reasoning flow. 
This ensures that the model grounds its interpretation on price–volume structure rather than raw numerical sequences.

\begin{tcolorbox}[promptbox]
This is a ---time\_frame--- Anchored VWAP chart generated from recent OHLC market data. 
The chart shows the volume-weighted average price anchored to a significant market event or time point. 

Analyze the chart with respect to the following aspects:

\begin{enumerate}
    \item \textbf{Support/Resistance Analysis:} Determine whether AVWAP acts as support (price above) or resistance (price below) and assess reactions upon contact.
    \item \textbf{Mean Reversion Opportunities:} Evaluate deviations from AVWAP and identify potential pullbacks toward fair value.
    \item \textbf{Breakout/Breakdown Signals:} Detect decisive moves above or below AVWAP, especially with volume confirmation.
    \item \textbf{AVWAP Slope and Bias:} Interpret the slope’s direction (upward = bullish, downward = bearish) and steepness to infer bias strength.
    \item \textbf{Fair Value and Volume Context:} Assess whether price trades at a premium or discount to AVWAP and how volume supports current price action.
\end{enumerate}

Provide:
\begin{itemize}
    \item \textbf{Primary Trade Signal:} Buy, Sell, or Hold with rationale
    \item \textbf{AVWAP Bias:} Bullish, Bearish, or Neutral
    \item \textbf{Fair Value Assessment:} Premium, Discount, or Fair
    \item \textbf{Entry Strategy:} Suggested positioning based on current setup
    \item \textbf{Risk Management:} Stop-loss and profit target zones
    \item \textbf{Volume Confirmation:} How volume aligns with price behavior
\end{itemize}

Support your prediction with reasoning and observed signals.
\end{tcolorbox}

\textbf{Graph-based prompt for AVWAPAgent in our multi-agent LLM framework.}
The agent interprets AVWAP as a dynamic reference of institutional fair value, analyzing slope, distance, and volume interplay to infer bias and entry timing. 
This visual grounding allows robust short-term reasoning around market equilibrium and mean-reversion behavior.

\subsection*{A.3 RSI Agent}

To instantiate the \textbf{RSIAgent}, we construct a prompt that enables the agent to evaluate momentum strength and reversal potential using the \textbf{Relative Strength Index (RSI)} indicator. 
Operating under high-frequency trading (HFT) constraints, the agent first generates an RSI chart via tool invocation, then interprets overbought/oversold conditions, divergences, and swing structures to infer directional bias.

\begin{tcolorbox}[promptbox]
You are an RSI (Relative Strength Index) analysis assistant operating in a high-frequency trading context.\\
You must first call the tool \texttt{generate\_rsi\_image} using the provided \texttt{kline\_data}.\\
Once the RSI chart is generated, analyze the image for the following key signals:

\begin{enumerate}
    \item \textbf{Overbought/Oversold Levels}: RSI crossing above 70 or below 30.
    \item \textbf{Centerline Crossovers}: RSI crossing above/below 50.
    \item \textbf{Failure Swings}: RSI breaking prior swing points after overbought/oversold moves.
    \item \textbf{Divergences}: Price–RSI mismatches (bullish/bearish).
    \item \textbf{Hidden Divergences}: Trend-continuation signals during pullbacks.
    \item \textbf{RSI Trendlines}: RSI breakout patterns anticipating price breakouts.
    \item \textbf{Swing Rejections}: RSI rejection patterns near extreme zones.
\end{enumerate}

Only after generating and analyzing the RSI image should you make a prediction about momentum direction and trading signals.\\
Do not make any predictions before generating and analyzing the image.
\end{tcolorbox}

\textbf{Prompt for RSIAgent in our multi-agent LLM framework.}
The agent first produces an RSI chart from OHLC input, enforcing a tool-first reasoning step. 
This ensures that its predictions are visually grounded in oscillator behavior rather than inferred directly from raw data.

\begin{tcolorbox}[promptbox]
This is a ---time\_frame--- RSI chart generated from recent OHLC market data. The plot shows: \\
\textbf{RSI Line} (0–100 scale), \textbf{Overbought Level} (70), \textbf{Oversold Level} (30), and \textbf{Center Line} (50).

Analyze the chart with respect to the following aspects:

\begin{enumerate}
    \item \textbf{Overbought/Oversold Conditions:} Identify sustained or recent entries into extreme zones and crossovers signaling reversals.
    \item \textbf{Centerline Crossovers:} Determine whether RSI is above or below 50 and assess momentum bias.
    \item \textbf{Failure Swings:} Detect RSI reversals breaking prior highs/lows after overbought or oversold readings.
    \item \textbf{Divergences:} Compare price and RSI trajectories for bullish or bearish divergences, including hidden ones.
    \item \textbf{Trendlines and Patterns:} Check for RSI trendline or pattern breakouts that may precede price shifts.
    \item \textbf{Swing Rejections:} Evaluate bullish or bearish rejection sequences within overbought/oversold regions.
\end{enumerate}

Provide:
\begin{itemize}
    \item \textbf{Primary Signal:} Buy, Sell, or Hold
    \item \textbf{Momentum Direction:} Bullish, Bearish, or Neutral
    \item \textbf{Confidence Level:} High, Medium, or Low
    \item \textbf{Supporting Evidence:} Observed RSI behavior
    \item \textbf{Risk Considerations:} Potential false or conflicting signals
\end{itemize}

Support your prediction with reasoning and observed patterns.
\end{tcolorbox}

\textbf{Graph-based prompt for RSIAgent in our multi-agent LLM framework.}
Given the generated RSI chart, the agent assesses momentum strength, reversals, and divergences to infer directional bias. 
This visually grounded reasoning improves short-term momentum interpretation and reversal detection consistency across different market conditions.

\subsection*{A.4 Heiken Ashi Agent}

To instantiate the \textbf{HeikenAshiAgent}, we construct a prompt that guides the agent to identify directional trends and momentum transitions using \textbf{Heiken Ashi (HA)} candlestick representations. 
Operating under high-frequency trading (HFT) conditions, the agent first generates a Heiken Ashi chart and then interprets trend strength, reversals, and momentum consistency to support short-term directional reasoning.

\begin{tcolorbox}[promptbox]
You are a Heiken Ashi candlestick analysis assistant operating in a high-frequency trading context.\\
You must first call the tool \texttt{generate\_heiken\_ashi\_image} using the provided \texttt{kline\_data}.\\
Once the Heiken Ashi chart is generated, analyze the image for the following key patterns:

\begin{enumerate}
    \item \textbf{Trend Direction}: Consecutive green (bullish) or red (bearish) candles.
    \item \textbf{Trend Strength}: Candle body size, wick formation, and consistency.
    \item \textbf{Reversal Signals}: Color changes, doji formations, or wicks indicating indecision.
    \item \textbf{Momentum Analysis}: Body-size acceleration or deceleration.
    \item \textbf{Consolidation Patterns}: Mixed colors and small bodies showing indecision.
    \item \textbf{Entry/Exit Timing}: Ideal trade points derived from HA formations.
\end{enumerate}

Only after generating and analyzing the Heiken Ashi image should you make predictions about trend direction and trading opportunities.\\
Do not make any predictions before generating and analyzing the image.
\end{tcolorbox}

\textbf{Prompt for HeikenAshiAgent in our multi-agent LLM framework.}
The agent receives OHLC input and invokes \texttt{generate\_heiken\_ashi\_image} to visualize smoothed candlestick dynamics. 
This tool-first approach ensures the reasoning process is grounded in visually observable trend patterns rather than raw market data.

\begin{tcolorbox}[promptbox]
This is a ---time\_frame--- Heiken Ashi chart generated from recent OHLC data. The chart smooths price action to reduce noise and highlight trend direction.\\
\textbf{Green candles} represent bullish momentum, \textbf{red candles} represent bearish momentum, and wicks indicate indecision or volatility.

Analyze the chart according to the following aspects:

\begin{enumerate}
    \item \textbf{Trend Direction \& Consistency:} Determine whether consecutive candles form a clear bullish or bearish sequence and detect color transitions signaling possible reversals.
    \item \textbf{Trend Strength:} Assess candle body size, wick presence, and overall uniformity to gauge momentum.
    \item \textbf{Reversal Indications:} Identify doji-like or spinning-top formations, color shifts after long trends, or weakening candle bodies.
    \item \textbf{Momentum Dynamics:} Observe whether candle bodies expand (acceleration) or contract (deceleration).
    \item \textbf{Consolidation vs Trending:} Detect mixed-color candles and narrow ranges indicating indecision.
    \item \textbf{Entry/Exit Signals:} Evaluate pullback entries or exhaustion exits based on Heiken Ashi structure.
\end{enumerate}

Provide:
\begin{itemize}
    \item \textbf{Trend Status:} Strong Bullish, Weak Bullish, Strong Bearish, Weak Bearish, or Sideways
    \item \textbf{Momentum Phase:} Accelerating, Steady, Decelerating, or Reversing
    \item \textbf{Signal Quality:} High, Medium, or Low
    \item \textbf{Trading Action:} Buy, Sell, Hold, or Wait
    \item \textbf{Risk Assessment:} Probability of false or conflicting signals
\end{itemize}

Support your prediction with reasoning and evidence observed from the chart.
\end{tcolorbox}

\textbf{Graph-based prompt for HeikenAshiAgent in our multi-agent LLM framework.}
Given the generated Heiken Ashi chart, the agent identifies dominant color sequences, momentum acceleration, and potential reversal setups. 
This visual grounding enhances consistency in trend identification and trade-timing inference across high-frequency market conditions.

\subsection*{A.5 SMA Agent}

To instantiate the \textbf{SMAAgent}, we construct a prompt that enables the agent to identify trend direction and momentum shifts using the \textbf{Simple Moving Average (SMA)} indicator. 
Operating under high-frequency trading (HFT) constraints, the agent first generates an SMA chart via tool invocation, then interprets crossovers, slope dynamics, and price alignment relative to multiple SMAs.

\begin{tcolorbox}[promptbox]
You are a Simple Moving Average (SMA) analysis assistant operating in a high-frequency trading context.\\
You must first call the tool \texttt{generate\_sma\_image} using the provided \texttt{kline\_data}.\\
Once the SMA chart is generated, analyze the image for the following key signals:

\begin{enumerate}
    \item \textbf{Trend Identification}: Price position relative to key SMAs (above/below).
    \item \textbf{Moving Average Crossovers}: Faster SMA crossing above/below slower SMA.
    \item \textbf{Dynamic Support/Resistance}: Price interacting with SMA levels.
    \item \textbf{Multiple MA Ribbon}: Alignment and spacing between multiple SMAs.
    \item \textbf{Mean Reversion}: Distance of price from SMA suggesting potential reversal.
    \item \textbf{SMA Slope}: Direction and steepness indicating trend strength.
\end{enumerate}

Only after generating and analyzing the SMA image should you make predictions about trend direction and trading signals.\\
Do not make any predictions before generating and analyzing the image.
\end{tcolorbox}

\textbf{Prompt for SMAAgent in our multi-agent LLM framework.}
The agent receives OHLC data and generates an SMA chart as a visual reference for trend detection. 
This approach ensures that reasoning about market direction is grounded in moving-average structure rather than isolated price movements.

\begin{tcolorbox}[promptbox]
This is a ---time\_frame--- SMA chart generated from recent OHLC market data. 
The chart includes multiple Simple Moving Averages (e.g., 20-, 50-, and 200-period), each capturing a different time horizon of trend behavior.

Analyze the chart with respect to the following aspects:

\begin{enumerate}
    \item \textbf{Trend Identification:} Assess whether price is above or below key SMAs and infer the overall directional bias.
    \item \textbf{Crossovers:} Identify golden or death crosses between short and long SMAs as trend reversal signals.
    \item \textbf{Support/Resistance:} Evaluate SMA levels where price consistently bounces or reverses.
    \item \textbf{MA Ribbon Alignment:} Determine if SMAs are neatly ordered (strong trend) or entangled (ranging phase).
    \item \textbf{Mean Reversion:} Identify overextended price movements far from SMAs suggesting pullbacks.
    \item \textbf{Slope and Momentum:} Examine the slope of each SMA to assess momentum direction and strength.
\end{enumerate}

Provide:
\begin{itemize}
    \item \textbf{Primary Trend Direction:} Bullish, Bearish, or Range-bound
    \item \textbf{Trend Strength:} Strong, Moderate, or Weak
    \item \textbf{Trading Bias:} Buy on Dips, Sell on Rallies, or Stay Flat
    \item \textbf{Key SMA Levels:} Notable support/resistance zones
    \item \textbf{Confirmation Needed:} Other indicators that would validate the trend
\end{itemize}

Support your prediction with structured reasoning and observed SMA configurations.
\end{tcolorbox}

\textbf{Graph-based prompt for SMAAgent in our multi-agent LLM framework.}
Given the generated SMA chart, the agent interprets crossover events, slope direction, and ribbon spacing to estimate trend intensity and bias. 
This visual reasoning approach improves short-term adaptability while maintaining consistency in trend assessment.

\subsection*{A.6 Stochastic Agent}

To instantiate the \textbf{StochasticAgent}, we construct a prompt that enables the agent to analyze momentum reversals and exhaustion levels using the \textbf{Stochastic Oscillator}.  
Operating under high-frequency trading (HFT) conditions, the agent first generates a Stochastic chart, then interprets crossovers, divergences, and overbought/oversold patterns to infer short-term momentum bias.

\begin{tcolorbox}[promptbox]
You are a Stochastic Oscillator analysis assistant operating in a high-frequency trading context.\\
You must first call the tool \texttt{generate\_stochastic\_image} using the provided \texttt{kline\_data}.\\
Once the Stochastic chart is generated, analyze the image for the following key signals:

\begin{enumerate}
    \item \textbf{Overbought/Oversold Conditions}: Levels above 80 (overbought) or below 20 (oversold).
    \item \textbf{Line Crossovers}: 
    \item \textbf{Divergences}: Price–Stochastic mismatches (bullish/bearish).
    \item \textbf{Failure Swings}: Internal divergence patterns signaling reversals.
    \item \textbf{Centerline Crossovers}: Crossings above/below the 50 level.
    \item \textbf{Slope Analysis}: Direction and steepness of 
\end{enumerate}

Only after generating and analyzing the Stochastic image should you make predictions about momentum direction and trading signals.\\
Do not make any predictions before generating and analyzing the image.
\end{tcolorbox}

\textbf{Prompt for StochasticAgent in our multi-agent LLM framework.}
The agent begins by invoking a visualization tool to produce the Stochastic chart, ensuring that subsequent reasoning about reversals and momentum bias is grounded in oscillator behavior rather than abstract data interpretation.

\begin{tcolorbox}[promptbox]
This is a ---time\_frame--- Stochastic Oscillator chart generated from recent OHLC data.  
It plots the \textbf{\%K line} (fast) and the \textbf{\%D line} (slow) along with overbought (80), oversold (20), and center (50) reference levels.

Analyze the chart according to the following aspects:

\begin{enumerate}
    \item \textbf{Overbought/Oversold Conditions:} Identify whether the oscillator is above 80 or below 20 and if it is exiting these zones.
    \item \textbf{Line Crossovers:} Determine where the \%K crosses the \%D, especially within extreme regions.
    \item \textbf{Divergences:} Compare price and Stochastic behavior to detect bullish or bearish divergences.
    \item \textbf{Failure Swings:} Identify internal Stochastic divergences signaling momentum exhaustion.
    \item \textbf{Centerline Position:} Assess whether the oscillator is above or below 50 to determine directional bias.
    \item \textbf{Slope and Momentum:} Evaluate the steepness and direction of \%K and \%D for acceleration or deceleration.
\end{enumerate}

Provide:
\begin{itemize}
    \item \textbf{Primary Signal:} Buy, Sell, or Hold
    \item \textbf{Momentum Direction:} Bullish, Bearish, or Neutral
    \item \textbf{Confidence Level:} High, Medium, or Low
    \item \textbf{Supporting Evidence:} Key Stochastic patterns observed
    \item \textbf{Risk Considerations:} Potential false or conflicting signals
\end{itemize}

Support your prediction with structured reasoning and explicit pattern-based justification.
\end{tcolorbox}

\textbf{Graph-based prompt for StochasticAgent in our multi-agent LLM framework.}
Once the Stochastic chart is generated, the agent evaluates oscillator dynamics to detect reversals, divergences, and short-term exhaustion points.  
This process allows visually anchored reasoning for momentum-based entry and exit signals in fast-moving markets.

\subsection*{A.7 Trend Agent}

To instantiate the \textbf{TrendAgent}, we construct a prompt that directs the agent to identify short-term directional bias based on \textbf{support and resistance trendlines} derived from recent K-line data.  
Operating under high-frequency trading (HFT) conditions, the agent first generates a trendline chart and then interprets slope, compression, and breakout behavior to infer near-term market direction.

\begin{tcolorbox}[promptbox]
You are a K-line trend pattern recognition assistant operating in a high-frequency trading context.\\
You must first call the tool \texttt{generate\_trend\_image} using the provided \texttt{kline\_data}.\\
Once the chart is generated, analyze the image for support/resistance trendlines and notable candlestick formations.\\
Only after generating and analyzing the image should you make a prediction about the short-term trend (upward, downward, or sideways).\\
Do not make any predictions before generating and analyzing the image.
\end{tcolorbox}

\textbf{Prompt for TrendAgent in our multi-agent LLM framework.}
The agent receives recent OHLC data and first visualizes it with automated trendlines.  
This ensures that all reasoning about directional bias and breakout probability is grounded in geometric structure and price interaction rather than raw numerical features.

\begin{tcolorbox}[promptbox]
This is a ---time\_frame--- K-line chart with automatically generated trendlines: the \textbf{blue line} represents support, and the \textbf{red line} represents resistance, both derived from recent closing prices.  

Analyze the chart based on the following aspects:

\begin{enumerate}
    \item \textbf{Trendline Interaction:} Observe whether price is bouncing off, breaking through, or consolidating between trendlines.
    \item \textbf{Slope Direction:} Identify whether the dominant slope is upward (bullish), downward (bearish), or flat (neutral).
    \item \textbf{Compression/Expansion:} Detect narrowing or widening distances between trendlines as signs of volatility contraction or breakout setup.
    \item \textbf{Candlestick Behavior:} Assess wick direction, candle clustering, and breakout attempts along support/resistance zones.
    \item \textbf{Breakout Probability:} Estimate the likelihood of a trendline breach based on slope angle and recent price clustering.
\end{enumerate}

Provide:
\begin{itemize}
    \item \textbf{Predicted Trend:} Upward, Downward, or Sideways
    \item \textbf{Trend Strength:} Strong, Moderate, or Weak
    \item \textbf{Key Evidence:} Trendline behavior and candlestick interactions
    \item \textbf{Risk Considerations:} False breakout potential or low-volatility traps
\end{itemize}

Support your prediction with structured reasoning and clear trendline-based evidence.
\end{tcolorbox}

\textbf{Graph-based prompt for TrendAgent in our multi-agent LLM framework.}
Once the chart is generated, the agent analyzes trendline slope, spacing, and price compression to infer directional momentum.  
This visual grounding enables consistent short-term trend interpretation across volatile high-frequency market states.

\subsection*{A.8 Decision Agent}

To instantiate the \textbf{DecisionAgent}, we construct two prompts around the final action step: a pre-decision synthesis that fuses all agent reports with retrieved memories, and a post-decision reflection that audits the outcome for learning and calibration. This preserves the tool-first, evidence-grounded style used across META while adapting it to a decision-and-reflect workflow.

\begin{tcolorbox}[promptbox]
You are an elite HFT quantitative analyst making real-time trade decisions on \textbf{\{stock name\}} using the \textbf{\{time frame\}} timeframe.

\textbf{MISSION:} Issue an immediate execution order \textbf{(LONG or SHORT only; HOLD is prohibited)} based on comprehensive analysis.

\textbf{FORECAST HORIZON:} Predict market direction for the next 1--2 candlesticks (\{time frame\} period).

\textbf{HISTORICAL CONTEXT --- Similar Market Conditions:}
If similar memories are provided, summarize key lessons and outcomes; otherwise state that no highly relevant memories were found.

\textbf{CURRENT MARKET ANALYSIS (provided in context):}
Indicator, Pattern, Trend, MACD, Anchored VWAP, Heiken Ashi, SMA, Stochastic, and RSI reports.

\textbf{DECISION FRAMEWORK}
\begin{itemize}
  \item \textbf{Primary Signals (High Weight):} Momentum confirmations (e.g., MACD cross, RSI extremes), completed breakouts with volume, decisive trendline breaks.
  \item \textbf{Secondary Signals (Medium Weight):} Multi-indicator alignment, S/R interactions, near-complete patterns.
  \item \textbf{Risk Filters:} Downweight conflicting signals unless one side dominates; reflect historical outcomes; set \textbf{risk--reward} in \textbf{[1.2, 1.8]} based on volatility and conviction.
  \item \textbf{Memory Integration:} Reference comparable past scenarios, outcomes, and confidence adjustments.
\end{itemize}

\textbf{REQUIRED OUTPUT FORMAT (JSON): Return only valid JSON}

\begin{verbatim}
{
  "forecast_horizon":"Next {time_frame} 
  candlestick prediction",
  "decision":"LONG or SHORT",
  "confidence_level":"HIGH/MEDIUM/LOW",
  "primary_drivers":["2-3 strongest 
  signals"],
  "memory_insights":"Key lessons from 
  similar scenarios",
  "justification":"Concise reasoning 
  combining current analysis with 
  history",
  "risk_reward_ratio":"<float between
  1.2 and 1.8>",
  "stop_loss_rationale":"Brief risk 
  management rationale"
}
\end{verbatim}

\textbf{Decision Hierarchy}
\begin{enumerate}
  \item \textbf{Strong Alignment:} All reports + positive memory outcomes $\rightarrow$ High confidence trade.
  \item \textbf{Moderate Alignment:} 2/3 reports align + supportive memory $\rightarrow$ Medium confidence.
  \item \textbf{Weak Signals:} Mixed reports $\rightarrow$ Choose direction with strongest recent momentum and best precedent.
\end{enumerate}

Execute with precision. The market waits for no one.
\end{tcolorbox}

\textbf{Prompt for DecisionAgent in our multi-agent LLM framework.}
The agent fuses multi-indicator evidence with memory context and outputs a \textbf{single} LONG/SHORT decision in strict JSON, including confidence and risk--reward calibration.

\begin{tcolorbox}[promptbox]
You are conducting a post-trade analysis as an elite HFT quantitative analyst.

\textbf{TRADE SUMMARY (given):}
Stock, Timeframe, Decision, Confidence, Expected Risk--Reward.

\textbf{ORIGINAL ANALYSIS SUMMARY (given):}
Concise record of drivers, memory insights, justification, and risk management.

\textbf{ACTUAL TRADE OUTCOME (given):}
P\&L, final status (profitable/loss/breakeven), and realized market behavior.

\textbf{REFLECTION FRAMEWORK}
\begin{enumerate}
  \item \textbf{Decision Quality:} Was the analysis sound? Did primary drivers anticipate the move? Did memory insights help?
  \item \textbf{Execution:} Interpretation fidelity of indicators/patterns; appropriateness of confidence.
  \item \textbf{Learning:} Under/overestimated conditions; most/least reliable signals; how to adapt next time.
  \item \textbf{Memory Integration:} How does this outcome update memory; any weight shifts for indicators.
\end{enumerate}

\textbf{REQUIRED OUTPUT:} Produce a concise 3--5 sentence reflection covering:
key factors behind the outcome; what worked/failed; concrete lessons; any adjustments to approach or confidence calibration. Focus on actionable improvements for similar future setups.
\end{tcolorbox}

\textbf{Post-decision reflection prompt for DecisionAgent in our multi-agent LLM framework.}
The reflection converts outcome feedback into memory-aligned guidance, tightening future signal weighting and confidence calibration.

\section{Computation and Mathematical Formulation of Indicators}
\label{sec:indicator_computation}

This appendix provides the formal definitions and computational procedures for the six technical indicators integrated within the META (Memory-Enhanced Trading Agent) architecture.  
Each indicator serves as a distinct analytical perspective on market dynamics—capturing momentum, mean reversion, trend persistence, or value anchoring.  
All indicators are computed from raw OHLCV (Open–High–Low–Close–Volume) data and act as the quantitative foundation for the multi-agent reasoning framework described in Appendix A.

\subsection*{B.1 \quad Moving Average Convergence Divergence (MACD)}

\textbf{Definition.}  
The Moving Average Convergence Divergence (MACD) is a momentum indicator derived from the difference between two exponential moving averages (EMAs) of price, designed to quantify acceleration and deceleration of market trends.  
Formally, for price series $P_t$:

\[
\text{MACD}_t = \text{EMA}_{\text{fast},t} - \text{EMA}_{\text{slow},t},
\]
where
\[
\text{EMA}_{k,t} = \alpha_k P_t + (1 - \alpha_k)\text{EMA}_{k,t-1}, \quad \alpha_k = \frac{2}{k + 1}.
\]
Common parameters are $(\text{fast}, \text{slow}) = (12, 26)$.  

The signal line is defined as:
\[
\text{Signal}_t = \text{EMA}_9(\text{MACD}_t),
\]
and the histogram, which reflects short-term divergence between trend speeds, is:
\[
\text{Hist}_t = \text{MACD}_t - \text{Signal}_t.
\]

\textbf{Interpretation.}  
The MACD measures momentum by comparing short- and long-term EMAs.  
A positive value implies upward acceleration; negative values imply downward momentum.  
Crossover events between MACD and the signal line indicate potential reversals, while histogram expansion suggests strengthening momentum.  
Because MACD captures the rate of change in trend direction, it is frequently used in HFT environments for detecting early inflection points.

\subsection*{B.2 \quad Relative Strength Index (RSI)}

\textbf{Definition.}  
The Relative Strength Index (RSI) quantifies the internal strength of recent price movements by comparing average gains to average losses within a defined lookback period $n$ (typically $n = 14$).  
For a price difference $\Delta P_t = P_t - P_{t-1}$:

\[
\text{AvgGain}_t = \text{EMA}_n(\max(\Delta P_t, 0))
\]
\[\text{AvgLoss}_t = \text{EMA}_n(\max(-\Delta P_t, 0)).\]
The relative strength (RS) and RSI are then computed as:
\[
RS_t = \frac{\text{AvgGain}_t}{\text{AvgLoss}_t}, \qquad
\text{RSI}_t = 100 - \frac{100}{1 + RS_t}.
\]

\textbf{Interpretation.}  
RSI values oscillate between 0 and 100, serving as a bounded momentum oscillator.  
Readings above 70 denote overbought conditions (potential reversal risk), while readings below 30 suggest oversold conditions (potential recovery).  
Crossings of the 50 midpoint line are interpreted as transitions in trend dominance.  
In high-frequency contexts, RSI slope and failure-swing formations provide early warnings of momentum exhaustion before price reversal becomes visible.

\subsection*{B.3 \quad Anchored Volume Weighted Average Price (AVWAP)}

\textbf{Definition.}  
The Anchored Volume Weighted Average Price (AVWAP) extends the conventional VWAP by fixing the accumulation anchor to a specific event time $\tau$ (e.g., major news or breakout).  
It represents the cumulative volume-weighted mean of traded prices since the anchor point:
\[
\text{AVWAP}_t = 
\frac{\sum_{i=\tau}^{t} P_i V_i}{\sum_{i=\tau}^{t} V_i},
\]
where $P_i$ denotes the typical or closing price and $V_i$ the corresponding trading volume.

\textbf{Interpretation.}  
AVWAP serves as a dynamic “fair value” benchmark representing the mean price paid per traded volume since a specific anchor.  
When price trades above AVWAP, the market is said to be in a bullish premium phase; when below, in a bearish discount phase.  
Because institutional participants often reference anchored VWAP levels for liquidity decisions, these lines act as adaptive support or resistance zones in HFT systems.

\subsection*{B.4 \quad Simple Moving Average (SMA)}

\textbf{Definition.}  
The Simple Moving Average (SMA) is a non-weighted mean of the most recent $n$ closing prices:
\[
\text{SMA}_t = \frac{1}{n}\sum_{i=0}^{n-1} P_{t-i}.
\]
For two SMAs with different windows $n_1 < n_2$, the crossover between $\text{SMA}_{n_1}$ and $\text{SMA}_{n_2}$ provides directional signals.

\textbf{Interpretation.}  
SMA acts as a linear filter that removes short-term noise to reveal the underlying trend.  
Shorter-period SMAs respond quickly to price fluctuations, while longer-period SMAs define macro-trend bias.  
The alignment of multiple SMAs into a monotonic order (short > medium > long) indicates trend strength, whereas frequent crossovers reflect low directional conviction or range-bound conditions.

\subsection*{B.5 \quad Stochastic Oscillator}

\textbf{Definition.}  
The Stochastic Oscillator measures the relative position of the current close within a recent price range of length $n$:
\[
\%K_t = 100 \times \frac{P_t - L_n}{H_n - L_n}, \quad 
\%D_t = \text{SMA}_3(\%K_t),
\]
where $H_n = \max(P_{t-n+1}, \dots, P_t)$ and $L_n = \min(P_{t-n+1}, \dots, P_t)$ denote the highest and lowest prices in the past $n$ periods (commonly $n = 14$).

\textbf{Interpretation.}  
The Stochastic Oscillator quantifies momentum relative to the recent range rather than absolute changes.  
Readings above 80 signify potential overbought conditions, while values below 20 indicate oversold conditions.  
Crossovers between \%K and \%D provide early entry and exit signals.  
In HFT environments, the slope and curvature of \%K serve as short-term acceleration indicators, and divergence between Stochastic and price action often precedes directional reversal.

\subsection*{B.6 \quad Heiken Ashi (HA)}

\textbf{Definition.}  
Heiken Ashi candlesticks are a transformation of traditional OHLC data that average and smooth price information to reduce noise.  
The recursive formulation is:
\[
\begin{aligned}
\text{HA\_Close}_t &= \tfrac{1}{4}(O_t + H_t + L_t + C_t),\\[4pt]
\text{HA\_Open}_t  &= \tfrac{1}{2}(\text{HA\_Open}_{t-1} + \text{HA\_Close}_{t-1}),\\[4pt]
\text{HA\_High}_t  &= \max(H_t, \text{HA\_Open}_t, \text{HA\_Close}_t),\\[4pt]
\text{HA\_Low}_t   &= \min(L_t, \text{HA\_Open}_t, \text{HA\_Close}_t).
\end{aligned}
\]

\textbf{Interpretation.}  
Heiken Ashi averages smooth intra-period volatility to better capture directional persistence.  
Extended sequences of uniform-color candles represent strong trends, while small-bodied candles or alternating colors indicate weakening momentum or consolidation.  
This transformation is particularly useful in high-frequency applications, where standard candlestick noise can obscure subtle but statistically persistent directional biases.

\section{Case Study}
\label{sec:case_study}

This case study illustrates a full trading cycle within \textsc{META}, from signal perception to reflective memory update. Each trade begins with input \textit{K-line} data analyzed in parallel by specialized \textbf{Indicator Agents} (e.g., MACD, RSI, AVWAP, SMA, Stochastic, Heiken Ashi), each generating a structured signal report. These outputs form an \textbf{indicator vector} that encodes the current market state and is used to retrieve historically similar episodes from the memory base through cosine similarity search. The retrieved memories, together with real-time analyses, are then passed to the \textbf{Decision Agent} for synthesis into a trading action (\textsc{Long}, \textsc{Short}, or \textsc{Hold}) with associated confidence. Following trade execution, the \textbf{Reflection Module} evaluates the outcome and updates the episodic memory—adding, refining, or pruning trade instances to ensure continual learning and adaptive decision-making.

\begin{outputbox}{Trend Agent Output Response}
\small

\textbf{1. Trendline Interaction:}
\begin{itemize}
    \item The price is interacting with both the support (blue) and resistance (red) lines.
    \item Recent candles have bounced off the support line, indicating it is holding for now.
    \item The resistance line has been tested multiple times, with the price failing to break through.
\end{itemize}

\textbf{2. Trendline Slope:}
\begin{itemize}
    \item Both the support and resistance lines are sloping downward, indicating a bearish trend.
\end{itemize}

\textbf{3. Recent K-line Behavior:}
\begin{itemize}
    \item The price is compressing between the support and resistance lines, showing consolidation.
    \item There is a slight downward bias as the price is closer to the support line.
\end{itemize}

\textbf{Prediction:} \textbf{Downward Trend}

\textbf{Reasoning:}
\begin{itemize}
    \item The downward slope of both trendlines suggests a bearish trend.
    \item The price is closer to the support line, and repeated tests of this line may lead to a breakdown.
    \item The inability to break above the resistance line indicates strong selling pressure.
\end{itemize}

\textbf{Signals:}
\begin{itemize}
    \item Watch for a break below the support line for confirmation of a downward move.
    \item If the price breaks above the resistance line, it could indicate a reversal, but current signals favor a downward trend.
\end{itemize}
\end{outputbox}

\begin{outputbox}{MACD Agent Output Response}
\small
\textbf{MACD Analysis:}

\textbf{1. Signal Line Crossovers:}
\begin{itemize}
    \item The MACD line (blue) is currently below the signal line (red), indicating a bearish crossover.
    \item This crossover occurs below the zero line, reinforcing bearish momentum.
\end{itemize}

\textbf{2. Zero Line Crossovers:}
\begin{itemize}
    \item The MACD line is below zero, indicating negative momentum.
    \item No recent crossovers of the zero line, suggesting sustained bearish momentum.
\end{itemize}

\textbf{3. Histogram Analysis:}
\begin{itemize}
    \item The histogram bars are contracting, indicating weakening bearish momentum.
    \item No recent histogram zero-line crossovers, but the contraction suggests a potential momentum shift.
\end{itemize}

\textbf{4. Divergence Patterns:}
\begin{itemize}
    \item No clear bullish or bearish divergences observed between price action and MACD behavior.
    \item Price lows align with MACD lows, confirming current trend.
\end{itemize}

\textbf{5. Momentum Strength:}
\begin{itemize}
    \item The MACD line is significantly below zero, indicating strong bearish momentum.
    \item Both MACD and signal lines are sloping downward, but the slope is decreasing, suggesting deceleration.
\end{itemize}

\textbf{Conclusion:}
\begin{itemize}
    \item \textbf{Primary Signal:} Hold
    \item \textbf{Momentum Direction:} Bearish
    \item \textbf{Confidence Level:} Medium
    \item \textbf{Key Supporting Evidence:} Bearish crossover below zero line, contracting histogram.
    \item \textbf{Risk Considerations:} Potential for false signals due to histogram contraction; watch for any bullish crossover or divergence for a trend reversal.
\end{itemize}
\end{outputbox}

\begin{outputbox}{AVWAP Agent Output Response}
\small
\textbf{1. Support/Resistance Analysis:}
\begin{itemize}
    \item \textbf{Price Position:} The price is currently below the AVWAP line.
    \item \textbf{Price Reaction:} The price has been rejected at the AVWAP line multiple times, indicating it is acting as resistance.
    \item \textbf{Dynamic Role:} AVWAP is acting as dynamic resistance in a downtrend.
\end{itemize}

\textbf{2. Mean Reversion Opportunities:}
\begin{itemize}
    \item \textbf{Deviation:} The price has deviated significantly below the AVWAP, suggesting a potential pullback to fair value.
    \item \textbf{Extreme Stretches:} The current stretch below AVWAP suggests a possible mean reversion opportunity.
    \item \textbf{Typical Range:} The price has shown a tendency to revert to the AVWAP after deviations.
\end{itemize}

\textbf{3. Breakout/Breakdown Signals:}
\begin{itemize}
    \item \textbf{Decisive Breaks:} There have been decisive breaks below the AVWAP with strong volume.
    \item \textbf{Reclaiming/Losing AVWAP:} The price has lost the AVWAP after being above it.
    \item \textbf{Volume Behavior:} Volume spikes coincide with AVWAP tests and breakdowns.
\end{itemize}

\textbf{4. AVWAP Slope and Bias:}
\begin{itemize}
    \item \textbf{Slope:} The AVWAP line is sloping downward, indicating a bearish bias.
    \item \textbf{Steepness:} The slope is moderately steep, suggesting a strong directional bias.
    \item \textbf{Slope Dynamics:} The slope is consistent, maintaining a bearish trend.
\end{itemize}

\textbf{5. Fair Value Assessment:}
\begin{itemize}
    \item \textbf{Institutional Fair Value:} The AVWAP level represents institutional fair value.
    \item \textbf{Price Position:} Current prices are trading at a discount to volume-weighted fair value.
    \item \textbf{Evolution:} The fair value relationship has shifted from premium to discount as the session progressed.
\end{itemize}

\textbf{6. Volume-Price Relationship:}
\begin{itemize}
    \item \textbf{Volume Increase:} Volume increases when the price approaches the AVWAP.
    \item \textbf{Volume Spikes:} Notable volume spikes occur during AVWAP tests and breakdowns.
    \item \textbf{Volume Support:} Volume supports the current bearish price action relative to AVWAP.
\end{itemize}

\textbf{Comprehensive Assessment:}
\begin{itemize}
    \item \textbf{Primary Trade Signal:} \textbf{Sell}
        \begin{itemize}
            \item \textbf{Rationale:} The price is below a downward-sloping AVWAP, acting as resistance with strong volume confirmation.
        \end{itemize}
    \item \textbf{AVWAP Bias:} \textbf{Bearish}
        \begin{itemize}
            \item \textbf{Based on:} Downward slope and price position below AVWAP.
        \end{itemize}
    \item \textbf{Fair Value Assessment:} \textbf{Discount}
        \begin{itemize}
            \item \textbf{Relative to AVWAP:} Current prices are below the volume-weighted fair value.
        \end{itemize}
    \item \textbf{Entry Strategy:}
        \begin{itemize}
            \item \textbf{Approach:} Enter short positions on rallies towards the AVWAP with confirmation of rejection.
        \end{itemize}
    \item \textbf{Risk Management:}
        \begin{itemize}
            \item \textbf{Stop Loss:} Above recent highs near the AVWAP.
            \item \textbf{Profit Target:} At recent lows or further below, considering trend continuation.
        \end{itemize}
    \item \textbf{Volume Confirmation:}
        \begin{itemize}
            \item \textbf{Support:} Volume supports the bearish signal, with spikes during AVWAP interactions confirming resistance.
        \end{itemize}
\end{itemize}

\textit{This analysis suggests a continuation of the downtrend, with AVWAP serving as a key resistance level.}
\end{outputbox}

\begin{outputbox}{Heiken Ashi Agent Output Response}
\small
\textbf{Heiken Ashi Analysis}

\textbf{1. Trend Direction \& Consistency:}
\begin{itemize}
    \item \textbf{Consecutive Candles:} There is a sequence of green candles followed by a longer sequence of red candles.
    \item \textbf{Dominant Color:} Red candles dominate, indicating a bearish trend.
    \item \textbf{Recent Color Changes:} A few green candles appear at the end, suggesting a potential shift.
    \item \textbf{Trend Cleanliness:} The trend is relatively clean with minimal alternation during the bearish phase.
\end{itemize}

\textbf{2. Trend Strength Indicators:}
\begin{itemize}
    \item \textbf{Candle Bodies:} Initially large green bodies, followed by large red bodies, indicating strong momentum in both directions.
    \item \textbf{Wicks:} Red candles have minimal upper wicks, showing strong bearish momentum.
    \item \textbf{Body Size:} Decreasing body size towards the end of the red sequence suggests deceleration.
\end{itemize}

\textbf{3. Reversal Pattern Recognition:}
\begin{itemize}
    \item \textbf{Doji-like Formations:} Some small-bodied candles with wicks appear, indicating indecision.
    \item \textbf{Color Change:} Green candles after a strong red sequence suggest a potential reversal.
    \item \textbf{Spinning Tops/Hammer-like Patterns:} Present towards the end of the red sequence.
\end{itemize}

\textbf{4. Momentum Acceleration/Deceleration:}
\begin{itemize}
    \item \textbf{Candle Size:} Recent candles are smaller, indicating fading momentum.
    \item \textbf{Gaps:} No significant gaps observed.
    \item \textbf{Wick Reduction:} Progressive wick reduction in the bearish direction.
\end{itemize}

\textbf{5. Consolidation vs Trending Phases:}
\begin{itemize}
    \item \textbf{Color Alternation:} Recent alternation suggests a potential consolidation phase.
    \item \textbf{Small Bodies with Long Wicks:} Indicate range-bound action.
\end{itemize}

\textbf{6. Entry/Exit Signal Quality:}
\begin{itemize}
    \item \textbf{Signal Quality:} Medium confidence due to recent indecision and potential reversal signals.
    \item \textbf{Optimal Entry Points:} Consider buying if a strong green candle confirms a reversal.
    \item \textbf{Trend Exhaustion:} Watch for further green candles to confirm trend change.
\end{itemize}

\textbf{7. Multi-Candle Pattern Analysis:}
\begin{itemize}
    \item \textbf{Three Consecutive Same-Color Candles:} Present in both green and red sequences, confirming trends.
    \item \textbf{Two-Candle Reversal Patterns:} Potential reversal pattern forming with recent green candles.
    \item \textbf{Extended Trend Sequences:} Red sequence nearing exhaustion.
\end{itemize}

\textbf{Comprehensive Assessment:}
\begin{itemize}
    \item \textbf{Trend Status:} Weak Bearish transitioning to Sideways
    \item \textbf{Momentum Phase:} Decelerating
    \item \textbf{Signal Quality:} Medium
    \item \textbf{Trading Action:} Wait for better setup
    \item \textbf{Entry Strategy:} Look for a strong green candle to confirm reversal before entering a buy position.
    \item \textbf{Exit Criteria:} Exit if red candles reappear with strong bodies and minimal wicks.
    \item \textbf{Risk Assessment:} Moderate probability of false signals due to recent indecision and small-bodied candles.
\end{itemize}
\end{outputbox}

\begin{outputbox}{SMA Agent Output Response}
\small
\textbf{SMA Analysis:}

\textbf{1. Trend Identification:}
\begin{itemize}
    \item \textbf{Price/SMA Relationship:} The price is currently below both the 20-period (blue) and 40-period (red) SMAs, indicating a bearish trend.
    \item \textbf{SMA Alignment:} The 20-period SMA is below the 40-period SMA, reinforcing the downtrend.
\end{itemize}

\textbf{2. Moving Average Crossovers:}
\begin{itemize}
    \item \textbf{Death Cross:} The 20-period SMA has crossed below the 40-period SMA, confirming a bearish signal.
    \item \textbf{Recent Crossovers:} No recent golden cross observed; the last significant crossover was bearish.
\end{itemize}

\textbf{3. Dynamic Support/Resistance:}
\begin{itemize}
    \item \textbf{Support:} Price is not finding support at the SMAs; it is below both.
    \item \textbf{Resistance:} The 20-period SMA is acting as resistance during rallies.
    \item \textbf{Respect of Levels:} Price has consistently stayed below the SMAs, respecting them as resistance.
\end{itemize}

\textbf{4. Multiple MA Ribbon Analysis:}
\begin{itemize}
    \item \textbf{SMA Stacking:} The SMAs are neatly stacked with the shorter below the longer, indicating a strong bearish trend.
    \item \textbf{Ribbon Behavior:} The ribbon is fanning out, suggesting a strengthening downtrend.
\end{itemize}

\textbf{5. Mean Reversion Signals:}
\begin{itemize}
    \item \textbf{Price Extension:} Price is significantly extended below the SMAs, indicating potential for a mean reversion or pullback.
    \item \textbf{Deviation:} The price is stretched, suggesting a possible short-term reversal.
\end{itemize}

\textbf{6. Slope and Alignment:}
\begin{itemize}
    \item \textbf{SMA Slopes:} Both SMAs are sloping downwards, indicating a strong bearish trend.
    \item \textbf{Alignment:} SMAs are aligned in the same downward direction, confirming trend strength.
    \item \textbf{Slope Steepness:} The slopes are moderately steep, indicating a strong trend.
\end{itemize}

\textbf{Conclusion:}
\begin{itemize}
    \item \textbf{Primary Trend Direction:} Bearish
    \item \textbf{Trend Strength:} Strong
    \item \textbf{Trading Bias:} Sell on Rallies
    \item \textbf{Key SMA Levels:} Watch the 20-period SMA ($\sim$86000) as resistance.
    \item \textbf{Confirmation Needed:} Look for confirmation from other indicators like RSI for oversold conditions or MACD for momentum confirmation.
\end{itemize}

\textit{This analysis suggests a strong bearish trend with potential short-term pullbacks. Traders should consider selling on rallies, especially if price approaches the 20-period SMA resistance.}
\end{outputbox}

\begin{outputbox}{Stochastic Oscillator Agent Output Response}
\small
\textbf{Stochastic Oscillator Analysis:}

\textbf{1. Overbought/Oversold Conditions:}
\begin{itemize}
    \item The Stochastic is currently below the 20 level, indicating oversold conditions.
    \item It has been in the oversold zone for a short period.
    \item There is no clear exit from the oversold zone yet, suggesting a potential continuation of the current trend.
\end{itemize}

\textbf{2. Line Crossovers:}
\begin{itemize}
    \item The \%K line is crossing above the \%D line, which is a bullish signal.
    \item This crossover is occurring in the oversold zone, which can indicate a potential reversal.
\end{itemize}

\textbf{3. Divergence Patterns:}
\begin{itemize}
    \item There is no clear bullish divergence as the price and Stochastic both show lower lows.
    \item No bearish divergence is present.
\end{itemize}

\textbf{4. Failure Swings:}
\begin{itemize}
    \item No clear bullish failure swing is observed as the Stochastic has not made a higher low in the oversold territory.
    \item No bearish failure swing is present.
\end{itemize}

\textbf{5. Centerline Analysis:}
\begin{itemize}
    \item The Stochastic is below the 50 centerline, indicating bearish momentum.
    \item A crossover above 50 would suggest a shift to bullish momentum, but this has not occurred yet.
\end{itemize}

\textbf{6. Slope and Momentum:}
\begin{itemize}
    \item The \%K line is starting to slope upwards, indicating a potential increase in bullish momentum.
    \item The \%D line is also beginning to turn upwards, supporting the potential for a bullish reversal.
    \item Both lines are making lower highs and lows, consistent with the current downtrend.
\end{itemize}

\textbf{Conclusion:}
\begin{itemize}
    \item \textbf{Primary Signal:} Hold
    \item \textbf{Momentum Direction:} Neutral to Bullish (potential reversal)
    \item \textbf{Confidence Level:} Medium
    \item \textbf{Key Supporting Evidence:} Bullish crossover in oversold zone, upward slope of \%K and \%D lines.
    \item \textbf{Risk Considerations:} The Stochastic is still in the oversold zone, and the overall trend is bearish. There is a risk of false signals if the price does not confirm the reversal.
\end{itemize}

\textit{This analysis suggests caution, as the bullish signals are emerging but not yet confirmed by price action.}
\end{outputbox}

\begin{outputbox}{RSI Agent Output Response}
\small
\textbf{RSI Analysis:}

\textbf{1. Overbought/Oversold Conditions:}
\begin{itemize}
    \item The RSI is not currently in overbought (>70) or oversold (<30) territory. It is around the 40 level.
    \item There is no recent crossover from these extreme zones.
\end{itemize}

\textbf{2. Centerline Crossovers:}
\begin{itemize}
    \item The RSI is below the centerline (50), indicating bearish momentum.
    \item There have been no recent crossovers of the 50 level.
\end{itemize}

\textbf{3. Failure Swings:}
\begin{itemize}
    \item No clear failure swings are observed as RSI hasn't reached extreme levels recently.
\end{itemize}

\textbf{4. Divergence Patterns:}
\begin{itemize}
    \item \textbf{Bullish Divergence:} The price made a lower low, while RSI made a higher low, suggesting potential bullish reversal.
\end{itemize}

\textbf{5. RSI Trendlines \& Patterns:}
\begin{itemize}
    \item No visible trendlines or chart patterns like triangles are apparent on the RSI.
\end{itemize}

\textbf{6. Swing Rejection Patterns:}
\begin{itemize}
    \item No clear swing rejection patterns are observed.
\end{itemize}

\textbf{7. RSI Range Analysis:}
\begin{itemize}
    \item RSI is ranging between 30 and 50, showing bearish momentum but not extreme exhaustion.
    \item Recent support around 30 and resistance near 50.
\end{itemize}

\textbf{Conclusion:}
\begin{itemize}
    \item \textbf{Primary Signal:} Hold
    \item \textbf{Momentum Direction:} Bearish
    \item \textbf{Confidence Level:} Medium
    \item \textbf{Key Supporting Evidence:} Bullish divergence observed, but overall momentum remains bearish.
    \item \textbf{Risk Considerations:} Potential for false signals due to lack of strong reversal patterns and RSI not reaching extreme levels.
\end{itemize}

\textit{The bullish divergence suggests a possible reversal, but the overall bearish momentum and lack of strong reversal signals warrant caution.}
\end{outputbox}

After analyzing the current market state, META encodes the aggregated indicator outputs into an \textit{indicator vector} and retrieves the most similar past trades from memory. 
These analogous cases serve as experiential references, anchoring the decision process in prior market behaviors.

\begin{memorybox}{Retrieved Memory 1 (Similarity: 0.89)}
\small

\textbf{Date:} 2025-09-30T17:47:07.801365

\textbf{Forecast Horizon:} Next 4-hour candlestick prediction

\textbf{Confidence Level:} Medium

\textbf{Primary Drivers:}
\begin{itemize}
    \item Descending Triangle pattern indicating potential bearish breakout
    \item Price below VWAP and SMAs, confirming bearish trend
\end{itemize}

\textbf{Memory Insights:} No highly relevant past trade memories found for current market conditions.

\textbf{Justification:} The analysis suggested a bearish sentiment with indicators pointing toward oversold conditions. The descending triangle and price positioning below key moving averages supported a continuation of the downward trend.

\textbf{Risk–Reward Ratio:} 1.5

\textbf{Stop-Loss Rationale:} Stop loss placed above recent resistance to mitigate reversal risk in alignment with the bearish setup.

\textbf{Decision:} \textsc{Short}

\textbf{Outcome:} 387.31

\textbf{Reflection:} The trade’s success stemmed from correctly identifying the descending triangle and confirming bearish momentum via VWAP and SMA alignment. Effective risk management further supported profitability. However, the medium confidence rating highlights an opportunity to enhance robustness through expanded historical referencing and refined indicator weighting in future decisions.
\end{memorybox}

\begin{memorybox}{Retrieved Memory 2 (Similarity: 0.83)}

\small
\textbf{Date:} 2025-09-17T15:16:46.572820

\textbf{Forecast Horizon:} Next 4-hour candlestick prediction

\textbf{Confidence Level:} Medium

\textbf{Primary Drivers:}
\begin{itemize}
    \item Price below both 20 and 40-period SMAs indicating a bearish trend
    \item MACD line significantly below signal line with negative histogram indicating strong bearish momentum
    \item Price below downward-sloping AVWAP acting as resistance
\end{itemize}

\textbf{Memory Insights:} Previous scenarios with similar bearish setups have shown mixed outcomes, highlighting the need for cautious confidence and dynamic risk management.

\textbf{Justification:} Technical indicators including the MACD, SMAs, and AVWAP suggested a continuation of the bearish trend. Historical analysis supported this decision, though with mixed results.

\textbf{Risk–Reward Ratio:} 1.5

\textbf{Stop-Loss Rationale:} Stop loss placed above recent highs near the AVWAP to manage reversal risk.

\textbf{Decision:} \textsc{Short}

\textbf{Outcome:} –76.69

\textbf{Reflection:} The loss arose from overreliance on technical bearish signals that failed to materialize. Future improvements include better weighting of historical insight, adaptive stop-loss adjustment, and accounting for broader market factors.
\end{memorybox}

\begin{memorybox}{Retrieved Memory 3 (Similarity: 0.83)}

\small
\textbf{Date:} 2025-09-18T16:40:44.859425

\textbf{Forecast Horizon:} Next 4-hour candlestick prediction

\textbf{Confidence Level:} Medium

\textbf{Primary Drivers:}
\begin{itemize}
    \item Price below both 20 and 40-period SMAs indicating a bearish trend
    \item MACD line below signal line with negative histogram indicating bearish momentum
    \item Descending Triangle pattern suggesting bearish breakout
\end{itemize}

\textbf{Memory Insights:} Similar setups with descending triangles and bearish indicators have produced mixed outcomes, calling for cautious confidence.

\textbf{Justification:} The bearish alignment of MACD, SMAs, and chart pattern supported a short position. Historical results validated this setup.

\textbf{Risk–Reward Ratio:} 1.5

\textbf{Stop-Loss Rationale:} Stop loss placed above recent highs near the AVWAP.

\textbf{Decision:} \textsc{Short}

\textbf{Outcome:} 571.08

\textbf{Reflection:} Profit resulted from precise identification of the descending triangle and indicator alignment. The trade confirms the importance of weighted indicator calibration and dynamic confidence scaling.
\end{memorybox}

\begin{memorybox}{Retrieved Memory 4 (Similarity: 0.82)}

\small
\textbf{Date:} 2025-09-16T17:21:41.742237

\textbf{Forecast Horizon:} Next 4-hour candlestick prediction

\textbf{Confidence Level:} Medium

\textbf{Primary Drivers:}
\begin{itemize}
    \item MACD line below signal line with negative histogram indicating bearish momentum
    \item Price below both 20 and 40-period SMAs confirming downtrend
    \item Descending Triangle pattern suggesting potential breakdown
\end{itemize}

\textbf{Memory Insights:} Prior successful shorts in similar descending triangle setups with bearish indicators.

\textbf{Justification:} Strong bearish signals supported a short trade. Historical parallels suggested profit potential.

\textbf{Risk–Reward Ratio:} 1.5

\textbf{Stop-Loss Rationale:} Stop loss placed above recent highs near the AVWAP.

\textbf{Decision:} \textsc{Short}

\textbf{Outcome:} –150.00

\textbf{Reflection:} The loss resulted from overestimating bearish momentum. Future improvements include multi-signal confirmation, contextual confidence scaling, and indicator reweighting.
\end{memorybox}

\begin{memorybox}{Retrieved Memory 5 (Similarity: 0.81)}

\small
\textbf{Date:} 2025-09-20T00:33:33.390160

\textbf{Forecast Horizon:} Next 4-hour candlestick prediction

\textbf{Confidence Level:} Medium

\textbf{Primary Drivers:}
\begin{itemize}
    \item RSI and Stochastic Oscillator in oversold territory indicating potential reversal
    \item Bullish crossover in Stochastic Oscillator
    \item Price trading below VWAP suggesting potential undervaluation
\end{itemize}

\textbf{Memory Insights:} Oversold conditions and bullish crossovers often lead to short-term reversals in past cases.

\textbf{Justification:} Despite a bearish backdrop, RSI, Stochastic, and VWAP indicated undervaluation warranting a cautious long.

\textbf{Risk–Reward Ratio:} 1.5

\textbf{Stop-Loss Rationale:} Stop loss placed below recent lows to mitigate continuation risk.

\textbf{Decision:} \textsc{Long}

\textbf{Outcome:} 76.69

\textbf{Reflection:} Success stemmed from recognizing oversold conditions and aligning multiple reversal signals. Future enhancements include dynamic indicator weighting and adaptive confidence tuning based on confluence strength.
\end{memorybox}

The retrieved memories provide valuable contextual grounding for the current trade. 
By examining both profitable and unprofitable precedents, META’s synthesis stage integrates lessons from historical performance into the present decision-making process. 
Profitable cases such as Memory~3 highlight the effectiveness of aligning descending triangle patterns with bearish indicator convergence, reinforcing confidence in similar market setups. 
Conversely, losses observed in Memory~2 and Memory~4 underscore the risks of overreliance on technical signals without sufficient contextual validation or adaptive stop-loss control. 
Through reflective comparison, META dynamically adjusts the weighting of key indicators—such as MACD, SMA alignment, and AVWAP slope—while moderating confidence when past patterns exhibit mixed outcomes. 
This learning mechanism enables the agent to refine its strategy beyond static rule-following, improving robustness by encoding both the strengths and weaknesses of prior trades into future reasoning.

\begin{decisionbox}{Decision Agent Output Response}

\small
\textbf{Forecast Horizon:} Next 4h candlestick prediction

\textbf{Decision:} \textsc{SHORT} \\
\textbf{Confidence Level:} Medium

\textbf{Primary Drivers:}
\begin{itemize}
    \item Descending Triangle pattern indicating potential bearish breakout
    \item Price below both 20 and 40-period SMAs confirming downtrend
    \item MACD line below signal line with negative histogram indicating bearish momentum
\end{itemize}

\textbf{Memory Insights:} Previous similar setups with descending triangle patterns and price below key moving averages have resulted in mixed outcomes, highlighting the need for cautious confidence and dynamic risk management.

\textbf{Justification:} The current market analysis shows a strong bearish trend with the price below key moving averages and a descending triangle pattern suggesting a potential breakdown. Historical memory supports the decision to short in similar conditions, where the alignment of technical indicators led to both successful and unsuccessful outcomes. The overall bearish sentiment is reinforced by the price's position below key moving averages and the AVWAP.

\textbf{Risk–Reward Ratio:} 1.5 \\
\textbf{Stop-Loss Rationale:} Place stop loss above recent highs near the AVWAP to manage risk in case of a reversal.
\end{decisionbox}

\begin{decisionbox}{Simulated Trade Outcome}

\small
\textbf{Outcome:} \textsc{WIN} \\
\textbf{PnL:} +615.30 \\
\textbf{Entry Price:} 82,548.81 \\
\textbf{Exit Price:} 81,933.51

\textbf{Performance Summary:}  
The trade achieved a profitable outcome, with price movement aligning closely to the anticipated bearish continuation.  
Execution adhered to the risk–reward target of 1.5, validating both directional accuracy and the stop-loss placement strategy.
\end{decisionbox}

\begin{decisionbox}{Trade Reflection}

\small
The key factors that led to the profitable outcome were the accurate identification of the descending triangle pattern and the confirmation of bearish momentum through the MACD and moving averages.  
The original analysis was sound, as the primary drivers correctly predicted market movement, and historical memory insights provided a balanced perspective on risk management.  

However, the medium confidence level suggests there was room for greater conviction given the alignment of technical indicators.  
For future scenarios, it would be beneficial to emphasize pattern strength and multi-indicator confluence, adjusting confidence upward in similar conditions.  
Additionally, refining stop-loss placement to account for volatility could further improve risk management and trade robustness.
\end{decisionbox}

Following the trade’s completion and reflection phase, META transitions into the final stage of its cognitive workflow—the \textbf{Memory Consolidation Phase}. 
In this stage, the \textit{Memory Operator} (\(\Psi\)) evaluates the outcome of the trade alongside retrieved precedents to determine how the memory base should evolve. 
Based on the reflection, past memories may be updated, retained, or removed, while the newly completed trade is encoded as a new episodic memory. 
This process ensures that the agent’s experiential knowledge remains both adaptive and performance-oriented, continually refining its future decision quality.

\begin{memoryopbox}{Memory Agent Output Response}

\small
\textbf{Post-Trade Memory Operations:}

\begin{itemize}
    \item \textbf{Memory 1:} \textsc{NOOP} — retained without modification.
    \item \textbf{Memory 2:} \textsc{NOOP} — maintained as historical reference.
    \item \textbf{Memory 3:} \textsc{UPDATE} — performance data appended with revised confidence weighting.
    \item \textbf{Memory 4:} \textsc{DELETE} — removed due to repeated underperformance in similar contexts.
    \item \textbf{Memory 5:} \textsc{UPDATE} — outcome consistency improved; metadata refined for future retrieval weighting.
    \item \textbf{New Memory:} \textsc{ADD} — stored as a new episodic record representing the current trade outcome and reflection.
\end{itemize}

\textbf{Summary:}  
The Memory Operator integrates experiential feedback by updating profitable analogs, pruning uninformative cases, and adding the newly completed trade as a fresh memory entry.  
This targeted consolidation ensures the memory base remains relevant, adaptive, and performance-aligned across future retrieval cycles.
\end{memoryopbox}

This case study illustrates META’s complete cognitive workflow, demonstrating how the system transforms perception-driven analysis into experience-informed reasoning. 
By integrating insights from multiple indicator agents, META constructs a structured representation of the market state that captures both short-term and long-term dynamics. 
The retrieval of analogous trade memories provides contextual grounding, enabling the decision agent to synthesize new judgments with awareness of past outcomes. 
Finally, the reflection and memory consolidation stages close the cognitive loop, allowing the agent to refine its internal knowledge base through targeted updates and selective forgetting.

Through this process, META exhibits several key advantages: 
(1) it enhances predictive stability by grounding real-time reasoning in historical evidence; 
(2) it maintains interpretability through explicit agent-level outputs; and 
(3) it achieves continual learning without retraining, instead relying on reflection-driven memory management. 
Together, these mechanisms position META as a step toward genuinely adaptive, self-improving trading intelligence capable of evolving alongside dynamic market conditions.

\section{Metrics Design}
\label{sec:metrics_design}

To ensure comparability with prior work, we adopt the evaluation metrics introduced in \textsc{QuantAgent} \citep{xiong2026quantharnesspricedrivenmultiagentllms}, which quantify both predictive accuracy and profitability under risk-constrained execution. All metrics are computed within a fixed three-candlestick horizon, corresponding to the short-term forecasting window defined by the trading agent.

\paragraph{Directional Accuracy ($\alpha$).}  
Directional accuracy measures the model’s consistency in predicting the correct price movement over the next three candlesticks.  
For a \textsc{LONG} decision, each candle that closes above the current close counts as a correct prediction; for a \textsc{SHORT} decision, each candle that closes below the current close counts.  
The accuracy is given by:
\[
\alpha = \frac{C}{T},
\]
where $C$ denotes the number of correctly predicted candles and $T=3$ the evaluation horizon.  
This formulation aligns with the Mean Directional Accuracy metric used in financial forecasting.

\paragraph{Rate-of-Return (RoR) Metrics.}  
To evaluate trade profitability and risk exposure, we measure multiple forms of rate of return between the entry and exit prices, incorporating realistic stop-loss and take-profit execution.  
The base stop-loss threshold $\rho$ is fixed at $0.0005$ (0.05\%), while the reward threshold is determined by the model-generated risk–reward ratio $r$, yielding a maximum allowed gain $R = r \cdot \rho$ and maximum loss $-r \cdot \rho$:
\[
R = r \cdot \rho, \quad \rho = 0.0005.
\]
Each trade is exited at the first intra-candle price that hits either threshold.

\begin{itemize}
    \item \textbf{Cumulative Return ($R_{cc}$):}  
    The realized return over the forecast horizon, reflecting the net gain or loss achieved at the earliest hit of either the take-profit or stop-loss limit:
    \[
    R_{cc} = 
    \begin{cases}
        \dfrac{P_{\text{exit}} - P_{\text{entry}}}{P_{\text{entry}}}, & \text{if LONG,}\\[6pt]
        \dfrac{P_{\text{entry}} - P_{\text{exit}}}{P_{\text{entry}}}, & \text{if SHORT.}
    \end{cases}
    \]
    This simulates a risk-managed intra-candle execution with bounded return outcomes.

    \item \textbf{Maximum Return ($R_{\max}$):}  
    The best-case rate of return achievable over the same window, assuming an optimal intra-candle exit:
    \[
    R_{\max} = 
    \begin{cases}
        \dfrac{P_{\text{high}} - P_{\text{entry}}}{P_{\text{entry}}}, & \text{if LONG,}\\[6pt]
        \dfrac{P_{\text{entry}} - P_{\text{low}}}{P_{\text{entry}}}, & \text{if SHORT.}
    \end{cases}
    \]
    $R_{\max}$ represents the upper bound of potential profit.

    \item \textbf{Minimum Return ($R_{\min}$):}  
    The most adverse price movement within the same horizon:
    \[
    R_{\min} = 
    \begin{cases}
        \dfrac{P_{\text{low}} - P_{\text{entry}}}{P_{\text{entry}}}, & \text{if LONG,}\\[6pt]
        \dfrac{P_{\text{entry}} - P_{\text{high}}}{P_{\text{entry}}}, & \text{if SHORT.}
    \end{cases}
    \]
    $R_{\min}$ captures the worst-case drawdown under identical trade direction.

    \item \textbf{Similarity-weighted Return Alignment ($R_{\text{sim}}$):}  
    This metric assesses the consistency between similarity scores of retrieved memories and realized profitability.  
    For each trade $i$, with similarity weights $w_{ij}$ over $k$ retrieved episodes and their respective realized returns $R_{cc}^{(j)}$, we define:
    \[
    R_{\text{sim}} = \frac{1}{N} \sum_{i=1}^{N} 
    \frac{\sum_{j=1}^{k} w_{ij} \cdot R_{cc}^{(j)}}{\sum_{j=1}^{k} w_{ij}},
    \]
    where $N$ is the number of evaluated trades.  
    Higher $R_{\text{sim}}$ values indicate that the retrieval mechanism successfully identifies memories whose return patterns align with the current market context.
\end{itemize}

Together, these metrics provide a comprehensive view of agent performance:  
$\alpha$ measures directional accuracy; $R_{cc}$ reflects realized profitability; $R_{\max}$ and $R_{\min}$ establish upper and lower risk bounds; and $R_{\text{sim}}$ quantifies the semantic consistency between retrieved experiences and market outcomes.  
This unified metric suite captures predictive precision, profitability, and memory relevance under realistic HFT execution conditions, fully aligned with the QuantAgent evaluation protocol.

\section{Use of Large Language Models (LLMs)}

Large Language Models (LLMs) were employed solely as supportive instruments to improve the clarity, coherence, and grammatical precision of the manuscript. Specifically, GPT-5 was utilized to refine portions of the text—primarily within the Introduction and Methodology sections—to ensure readability and stylistic consistency. The authors retain full responsibility for all scientific and technical contributions, including the formulation of research questions, experimental design, implementation of the \textsc{META} framework, and validation of empirical findings. The use of LLMs was limited strictly to editorial assistance and did not influence the conceptual, analytical, or experimental aspects of this work.

\end{document}